\documentclass[accepted]{uai2026} 
                        
\usepackage[american]{babel}

\usepackage{natbib} 
\usepackage{mathtools} 
\usepackage{booktabs} 
\usepackage{enumitem} 
\usepackage{tikz} 
\usetikzlibrary{calc,arrows.meta,positioning}
\usepackage{amsthm}
\usepackage{amssymb}
\usepackage[ruled,vlined]{algorithm2e}
\usepackage{pgfplots}
\usepackage{subcaption}
\pgfplotsset{compat=1.18}

\newtheorem{theorem}{Theorem}

\newtheorem{lemma}{Lemma}
\newtheorem{proposition}{Proposition}
\newtheorem{assumption}{Assumption}
\newtheorem{corollary}{Corollary}

\newcommand{\R}{\mathbb{R}}
\newcommand{\E}{\mathbb{E}}
\newcommand{\1}{\mathbf{1}}
\DeclareMathOperator{\Var}{Var}
\DeclareMathOperator{\Cov}{Cov}

\DeclareMathOperator{\VaR}{VaR}
\DeclareMathOperator{\ES}{ES}

\title{Diachronic Sample Integration: Robust Tail-Risk Estimation with Generative Models}

\author[1]{\href{mailto:<zsn20@mails.tsinghua.edu.cn>?Subject=Tail-Risk Estimation from Diffusion Models via Trajectory-Level Sample Integration}{Shuning~Zhao}{}}
\author[2]{\href{mailto:<patrick.wong@monash.edu>?Subject=Tail-Risk Estimation from Diffusion Models via Trajectory-Level Sample Integration}{Patrick~Wong}{}}
\author[3]{\href{mailto:<leran.zhang.1@student.unimelb.edu.au>?Subject=Tail-Risk Estimation from Diffusion Models via Trajectory-Level Sample Integration}{Leran~Zhang}{}}
\author[1]{\href{mailto:<xlhu@tsinghua.edu.cn>?Subject=Tail-Risk Estimation from Diffusion Models via Trajectory-Level Sample Integration}{Xiaolin~Hu}{}}
\affil[1]{%
    Department of Computer Science and Technology, BNRist\\
    Tsinghua University\\
    China
}
\affil[2]{%
    Department of Econometrics and Business Statistics\\
    Monash Business School\\
    Monash University\\
    Australia
}
\affil[3]{%
    School of Mathematics and Statistics\\
    University of Melbourne\\
    Australia
  }
  
\begin{document}
\maketitle

\begin{abstract}
Deep generative models are increasingly used as simulators for downstream decision making under data scarcity, but in risk-sensitive applications their usefulness depends on rare adverse scenarios rather than typical samples. Standard generative objectives prioritize bulk distributional fidelity, leaving low-probability tails vulnerable to localized optimization noise and making tail-dependent functionals unstable under finite simulation budgets. We introduce Diachronic Sample Integration (DSI), a test-time inference framework that ensembles generated samples across checkpoints from a stochastic training trajectory.
DSI targets a checkpoint-mixture distribution that averages checkpoint-specific tail fluctuations rather than relying on a single brittle endpoint.
We formalize this mechanism through a finite-budget bias-variance theory.
Empirically, across multivariate synthetic processes and high-frequency trading data, DSI substantially reduces tail-estimation error compared to single-checkpoint baselines under fixed simulation budgets, outperforming standard diffusion and state-of-the-art tail-aware baselines without modifying the generative objective.
\end{abstract}

\begin{figure*}
  \centering

\begin{tikzpicture}[
    font=\sffamily,
    panel/.style={rounded corners=2mm, draw=black!70, line width=0.6pt, fill=black!2},
    title/.style={font=\sffamily\small}, 
    small/.style={font=\sffamily\footnotesize},
    tiny/.style={font=\sffamily\tiny},
    arrow/.style={-{Latex[length=2.0mm]}, line width=0.7pt, draw=black!70},
    dot/.style={circle, fill=black!70, draw=none, inner sep=1.2pt}
]

\def\W{5.4}    
\def\H{3.75}   
\def\G{1.0}    

\coordinate (P1) at (0,0);
\coordinate (P2) at ({\W+\G},0);

\foreach \P/\Name in {P1/A, P2/C}{
  \coordinate (\Name-SW) at (\P);
  \coordinate (\Name-NE) at ($(\P)+(\W,\H)$);
  \draw[panel] (\Name-SW) rectangle (\Name-NE);
}

\node[title, anchor=north] at ($(A-SW)+(\W/2,{\H-0.15})$) {(a) Standard single checkpoint};
\node[title, anchor=north] at ($(C-SW)+(\W/2,{\H-0.15})$) {(b) DSI (Trajectory mixture)};

\begin{scope}
  \clip (A-SW) rectangle (A-NE);
  \def\base{0.4}

  \path[fill=blue!5] 
    ($(A-SW)+(0.3,\base)$) -- 
    plot[smooth] coordinates {
      ($(A-SW)+(0.3,\base)$) 
      ($(A-SW)+(1.3,0.7)$) 
      ($(A-SW)+(2.6,2.6)$) 
      ($(A-SW)+(3.9,0.7)$) 
      ($(A-SW)+(4.9,\base)$)
    } -- ($(A-SW)+(4.9,\base)$) -- cycle;
    
  \draw[line width=0.8pt, draw=blue!30, dashed] 
    plot[smooth] coordinates {
      ($(A-SW)+(0.3,\base)$) 
      ($(A-SW)+(1.3,0.7)$) 
      ($(A-SW)+(2.6,2.6)$) 
      ($(A-SW)+(3.9,0.7)$) 
      ($(A-SW)+(4.9,\base)$)
    };
  \node[small, text=blue!40] at ($(A-SW)+(4.3,1.4)$) {True};

  \path[fill=red!15, opacity=0.8] 
    ($(A-SW)+(1.0,\base)$) -- 
    plot[smooth] coordinates {
      ($(A-SW)+(1.0,\base)$) 
      ($(A-SW)+(1.8,0.9)$) 
      ($(A-SW)+(2.6,2.6)$) 
      ($(A-SW)+(3.4,0.9)$) 
      ($(A-SW)+(4.2,\base)$)
    } -- ($(A-SW)+(4.2,\base)$) -- cycle;

  \draw[line width=1.1pt, draw=red!70] 
    plot[smooth] coordinates {
      ($(A-SW)+(1.0,\base)$) 
      ($(A-SW)+(1.8,0.9)$) 
      ($(A-SW)+(2.6,2.6)$) 
      ($(A-SW)+(3.4,0.9)$) 
      ($(A-SW)+(4.2,\base)$)
    };

  \node[small, text=red!80, anchor=center] at ($(A-SW)+(2.6,0.6)$) {Generated};

  \draw[arrow, {Stealth[length=1.5mm]}-] ($(A-SW)+(1.1,\base+0.1)$) -- ++(-0.4,0.4)
    node[small, anchor=south, align=center, xshift=5pt] {Incorrect\\tails};
\end{scope}

\begin{scope}
  \clip (C-SW) rectangle (C-NE);
  \def\base{0.4}

  \path[fill=blue!5] 
    ($(C-SW)+(0.3,\base)$) -- 
    plot[smooth] coordinates {
      ($(C-SW)+(0.3,\base)$) 
      ($(C-SW)+(1.3,0.7)$) 
      ($(C-SW)+(2.6,2.6)$) 
      ($(C-SW)+(3.9,0.7)$) 
      ($(C-SW)+(4.9,\base)$)
    } -- ($(C-SW)+(4.9,\base)$) -- cycle;

  \draw[line width=0.8pt, draw=blue!30, dashed] 
    plot[smooth] coordinates {
      ($(C-SW)+(0.3,\base)$) 
      ($(C-SW)+(1.3,0.7)$) 
      ($(C-SW)+(2.6,2.6)$) 
      ($(C-SW)+(3.9,0.7)$) 
      ($(C-SW)+(4.9,\base)$)
    };
  \node[small, text=blue!40, anchor=west] at ($(C-SW)+(4.1,1.1)$) {True};

  \foreach \off/\h in {-0.3/-0.2, -0.1/-0.1, 0.2/0.3, 0.1/0.1}{
    \draw[line width=0.7pt, draw=red!40]
      plot[smooth, tension=0.7] coordinates {
        ($(C-SW)+(0.3, \base+\off)$)
        ($(C-SW)+(1.3,0.7+\h)$) 
        ($(C-SW)+(2.6, 2.6)$)
        ($(C-SW)+(3.9,0.7+\h)$) 
        ($(C-SW)+(4.9, \base+\off)$)
      };
  }

  \node[small, text=red!80, align=center, anchor=center] at ($(C-SW)+(2.6,0.9)$) {Model\\Checkpoints};

  \draw[arrow, {Stealth[length=1.5mm]}-] ($(C-SW)+(0.6,\base+0.1)$) -- ++(0.4,0.7) 
    node[small, anchor=south, align=center, xshift=-3pt] {Reduces tail\\fluctuations};
\end{scope}

\end{tikzpicture}
  \vspace{-2mm} 
  \caption{Conceptual illustration of DSI.
  (a) Single checkpoints prioritize bulk distributional fidelity, often truncating extreme tails due to optimization noise.
  (b) DSI aggregates heterogeneous checkpoints across a single trajectory, the resulting mixture averages checkpoint-specific tail fluctuations while any persistent trajectory bias remains as an irreducible floor.}
  \label{fig:dsi_intuition}
\end{figure*}
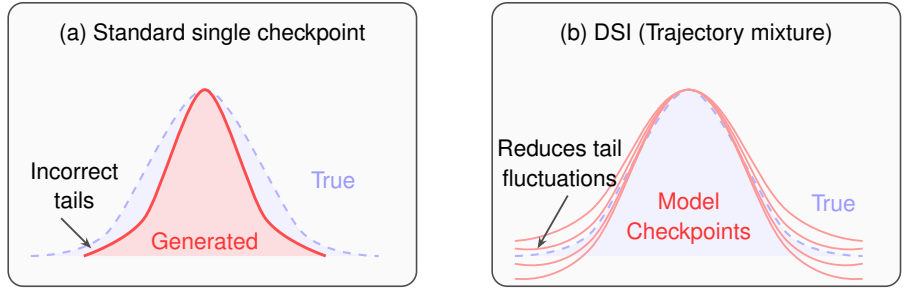

\section{Introduction}
\label{sec:intro}

Deep generative models are increasingly used as simulators for downstream decision making under data scarcity. In risk-sensitive applications, however, simulator reliability depends not only on typical samples but also on rare adverse scenarios. Reliable estimation of statistical functionals defined on low-probability regions is central to risk-sensitive inference. Because empirical tail observations are exceptionally scarce, practitioners increasingly rely on deep generative models to act as approximate simulators for downstream risk estimation \citep{gonen2025time,Tanaka2025,li2025mars}.\footnote{The source code for this project is available at https://github.com/ShuningZhao/DSI}

However, standard generative objectives, such as score matching or maximum likelihood, prioritize the ``bulk'' of the probability mass where data is dense. The extreme tails are consequently starved of reliable gradient signals and become dominated by localized optimization noise. Drawing synthetic samples from a single ``converged'' checkpoint often yields severe tail miscalibration, either arbitrarily truncating the tail (mode collapse) or hallucinating unrealistic extremes. Existing approaches attempt to resolve this during training via adversarial tail constraints or extreme-value reweighting \citep{Huang2024,galib2024fide}. While improving targeted metrics, these modeling interventions can alter the simulator's effective target distribution or impose additional structural assumptions.

DSI addresses a different practical problem: finite-budget risk estimates without imposing any parametric tail form, distributional constraint, or additional training objective. In many risk-management settings, the immediate task is not to recover an asymptotic tail law, but to estimate risk measures such as Value-at-Risk (VaR) and Expected Shortfall (ES) under a limited simulation budget constrained by compute, latency, or validation resources. We therefore propose Diachronic Sample Integration (DSI), a test-time framework that ensembles samples across stochastic training checkpoints (Figure~\ref{fig:dsi_intuition}). Rather than treating a single converged model state as the sole simulator, DSI uses the training trajectory as a source of checkpoint-level variability and aggregates generated samples to reduce tail-estimation instability while leaving the underlying generator unchanged.

DSI is not guaranteed to remove systematic model bias. Its statistical benefit arises when post-burn-in checkpoint errors contain averageable fluctuations around a persistent trajectory bias. We therefore analyze DSI through a finite-budget bias-variance decomposition for VaR and ES, showing that checkpoint stride and mixture size enter through an effective checkpoint count determined by tail-error autocorrelation. Empirically, on controlled synthetic processes and high-frequency NASDAQ data, DSI substantially reduces relative tail-estimation error compared to standard single-checkpoint diffusion models under matched sampling budgets, outperforming prior state-of-the-art tail-aware benchmarks.

This paper makes the following contributions:
\begin{itemize}[leftmargin=*,noitemsep]
    \item We introduce DSI, a lightweight test-time estimator that improves tail-risk inference via checkpoint sample ensembling.
    \item We formalize a bias-variance Trajectory Averaging Principle for VaR and ES under non-i.i.d.\ stratified samples, showing when checkpoint fluctuations can be averaged and when persistent bias remains.
    \item We show that DSI substantially reduces tail-estimation error and estimator variance under fixed simulation budgets, without retraining or changing the generative objective.
\end{itemize}

\section{Related Work}
\label{sec:related_work}





\subsection{Generative Simulators and Tail-Sensitive Objectives}

Deep generative models are increasingly employed as probabilistic simulators to mitigate data scarcity in risk-sensitive inference. Standard architectures \citep{Takahashi2025,Yuan2024,xia2024market,Yoon2019} optimize global distributional fidelity, which can leave low-probability regions vulnerable to localized optimization noise \citep{liu2024retrieval,kollovieh2023predict}.

Recent efforts to improve synthetic tail calibration have introduced adversarial risk penalties, extreme-aware diffusion mechanisms, or parametric extreme-value constraints \citep{Cont2025,galib2024fide,Huang2024}. Recent theory further shows that heavy-tailed target distributions pose distinct statistical challenges for score-based diffusion models: under polynomial tail decay, score-estimation and sampling rates explicitly depend on the tail index, unlike the exponentially decaying regime \citep{YuYu2026HeavyTailed}. A related but different line studies class-imbalanced long-tailed generation, where tail labels are underrepresented. For example, \citet{Das2025LongTailed} propose a training-time mutual-learning regularizer for conditional diffusion models to improve worst-case class-label performance.

These approaches address tail behavior by modifying the training objective, the modeling assumptions, or the conditional learning problem. In contrast, DSI preserves the original generative objective and addresses checkpoint-level tail instability at inference time via sample-space aggregation and trajectory-level uncertainty averaging.

\subsection{Model Ensembling and Averaging}

Ensembling methods such as Deep Ensembles, Stochastic Weight Averaging (SWA), and snapshot ensembles aggregate parameters or predictions to improve predictive performance and generalization \citep{Garipov2018, izmailov2018averaging, Maddox2019, huang2017snapshot, wortsman2022model}. These approaches operate in parameter or predictor space by averaging network weights (which can destroy complex learned manifolds) or point-predictions.

While these methods leverage multiple independent runs or training trajectories to find flatter optima, DSI shifts the focus to the sample space. By pooling stochastic samples across checkpoints, DSI constructs a macroscopic mixture distribution to stabilize non-linear statistical functionals. To address the high correlation often found between consecutive iterates in a single trajectory, we treat checkpoint spacing as a design variable via a temporal stride. This mechanism is designed to increase the effective checkpoint count by reducing autocorrelation among tail errors, a structural hypothesis we validate empirically in Section~\ref{sec:results}.


\section{Problem Formulation}
\label{sec:problem}

\subsection{Probability Space and Tail-Dependent Functionals}

We now formalize the downstream inference problem induced by a learned generative simulator: the simulator produces synthetic paths, which are mapped to a scalar decision quantity whose lower tail determines risk.

To formally ground the problem of tail estimation, let $(\Omega, \mathcal{F}, \mathbb{P})$ be a probability space, and let $X: \Omega \to \mathbb{R}$ be a random variable representing a scalar quantity of interest. In risk-sensitive applications, $X$ typically models the future return of a financial portfolio or the profit-and-loss (PnL) of a trading strategy, where large negative realizations correspond to severe and consequential losses. We denote the true, unknown data-generating distribution of $X$ by $P^\star$.

We are interested in estimating a statistical functional $\psi : \mathcal{P} \to \mathbb{R}$. Conceptually, $\psi$ is a mapping that takes an entire probability distribution as its input and outputs a single real-valued scalar (such as a risk metric). Here, $\mathcal{P}$ explicitly denotes the \emph{domain} of this functional, the class of admissible probability distributions over $\mathbb{R}$ for which $\psi$ is mathematically well-defined. For instance, evaluating Expected Shortfall strictly requires $\mathcal{P}$ to be restricted to the family of distributions possessing a finite first moment. We assume throughout that both the true target distribution $P^\star$ and our learned generative approximations belong to $\mathcal{P}$.

Throughout this work, we use Value-at-Risk (VaR) and Expected Shortfall (ES) as canonical examples of such functionals \citep{Artzner1999,rockafellar2000optimization}. 

For a given tail probability level $\alpha \in (0,1)$ (typically small, e.g., $\alpha=0.01$ or $0.05$), the $\alpha$-quantile, or Value-at-Risk, of a distribution $P$ with cumulative distribution function (CDF) $F_P$ is defined as:
\begin{equation}
q_\alpha(P) := \inf\{x \in \mathbb{R} : F_P(x) \ge \alpha\}.
\end{equation}
We use $\VaR_\alpha(P)$ and $q_\alpha(P)$ interchangeably for this lower-tail quantile.

\textbf{Intuition for VaR:} In risk management, Value-at-Risk (VaR) provides a severity threshold at a specified confidence level. For example, a 5\% lower-tail VaR identifies a loss threshold exceeded in approximately the worst 5\% of scenarios. VaR has historically been central to the Basel international banking regulatory framework, which sets global standards for bank capital and market-risk measurement \cite{BaselII}. However, VaR is merely a boundary: it is blind to the shape of the distribution beyond that threshold. It does not quantify how much the distribution can deteriorate in the extreme tail, nor the magnitude of losses once the VaR threshold is breached.

To capture the actual severity of extreme events, we use Expected Shortfall (ES) at level $\alpha$ in its lower-tail CDF form:
\begin{equation}
\ES_\alpha(P) := q_\alpha(P)-\frac{1}{\alpha}\int_{-\infty}^{q_\alpha(P)} F_P(x)\,dx,
\end{equation}
whenever the integral is finite. If $Y\sim P$ and $P$ is continuous with $F_P(q_\alpha(P))=\alpha$, this equals $\mathbb{E}_P[Y \mid Y \le q_\alpha(P)]$; for empirical distributions we use the corresponding fractional-order-statistic version. 

\textbf{Intuition for ES:} Expected Shortfall (ES) measures the average severity of outcomes in the extreme $\alpha$-tail. While VaR tells us the threshold of a ``bad day,'' ES tells us the average severity of losses on those bad days. This distinction is operationally important: the Basel III/FRTB market-risk framework \cite{BaselMarketRisk2019} moved from VaR toward ES as the headline internal-model risk measure because ES captures tail severity beyond the VaR cutoff. Because ES depends on the entire tail rather than a single point, it provides a more comprehensive measure of risk, but is significantly more difficult to estimate reliably from limited empirical data.

In many modern settings, direct empirical estimation of $\psi(P^\star)$ is infeasible due to the sheer scarcity of tail observations. Instead, practitioners increasingly rely on synthetic samples generated by a learned model $P_\theta$ to support downstream risk inference.

\subsection{Generative Approximation and the Single-Checkpoint Bottleneck}

Let $\Theta$ denote the parameter space of the generative simulator, and let $\mathcal{M} = \{P_\theta : \theta \in \Theta\}\subseteq\mathcal{P}$ denote the corresponding parametric class of generative distributions. We assume a mapping $\theta \mapsto P_\theta$ and fit the simulator by minimizing a global objective $\mathcal{L}(\theta; \mathcal{D})$ over a training dataset $\mathcal{D}$. Stochastic optimization traverses a complex loss landscape, inducing a training trajectory of model parameters $\theta_1, \theta_2, \dots, \theta_T$, where each iterate $\theta_t$ defines a specific generative distribution $P_t := P_{\theta_t}$.

Standard practice selects a single ``converged'' checkpoint (typically $\theta_T$) and draws a synthetic dataset $X_1,\dots,X_N \sim P_T$. This yields the plug-in estimator $\hat{\psi}_{T} := \psi(\hat P_T)$, where $\hat P_T$ is the empirical distribution of the generated samples.

However, relying on a single checkpoint is structurally brittle for tail estimation. Because global generative objectives (such as maximum likelihood or score matching) prioritize the ``bulk'' of the probability mass where data is dense, the tail regions of $P_t$ receive very little reliable gradient signal. Consequently, the tail behavior is highly sensitive to the optimization noise of the final training steps. A single checkpoint might arbitrarily truncate the tail (mode collapse) or hallucinate unrealistic extremes, leading to highly unreliable VaR and ES estimates.

\subsection{Checkpoint Ensembling on Generated Samples}
\label{subsec:ckp_ensembling}
To overcome the instability of single-checkpoint estimation, we treat the stochastic training trajectory not as a path to a single point estimate, but as a rich ensemble of tail approximations. Our core premise is that checkpoint ensembling on the generated samples leads to substantially better and more stable tail estimation.

Rather than discarding intermediate checkpoints, we aggregate samples drawn across them. Let $S \subset \{1,\dots,T\}$ be a selected set of checkpoints, and let $n_t$ denote the number of samples drawn from each checkpoint $P_t$, subject to a fixed total sampling budget $\sum_{t \in S} n_t = N$. 
Defining weights $w_t := n_t / N$, pooling these generated samples naturally targets the trajectory-induced mixture distribution.
For the theoretical analysis and experiments we focus on the equal-allocation case $w_t=1/K_S$, corresponding to Algorithm~\ref{alg:dsi} in the Appendix, but the framework is general and can accommodate any allocation scheme. The resulting mixture distribution is
\begin{equation}
\bar P_S = \sum_{t \in S} w_t P_t.
\end{equation}
The Diachronic Sample Integration (DSI) estimator is computed directly on the ensembled samples:
\begin{equation}
\hat{\psi}_{\mathrm{DSI}} := \psi(\hat P_S),
\end{equation}
where $\hat P_S$ is the empirical distribution of the pooled samples. 

\textbf{Why Ensembling Improves Tail Estimation:} By ensembling generated samples across the temporal trajectory, DSI averages checkpoint-specific tail errors that fluctuate across training. An isolated tail failure at one checkpoint can be diluted by other checkpoints, provided those failures are not perfectly correlated and do not share the same systematic direction. The mixture distribution $\bar P_S$ therefore reduces fluctuation-driven estimator variance while leaving persistent trajectory bias as an irreducible floor, all without modifying the underlying generative architecture or requiring additional training runs.

\subsection{Checkpoint Subsampling and Stride}
\label{subsec:ckp_stride}
Consecutive checkpoints along a stochastic training trajectory often induce highly correlated distributions. Pooling many nearby checkpoints may therefore introduce redundancy without significantly diversifying the simulated tail events.

To control correlation and maximize the informational value of the mixture, we parameterize $S$ using a stride $M \ge 1$ and a checkpoint count $K$:
\begin{equation*}
S(M,K) = \{t_0,\; t_0+M,\; t_0+2M,\;\dots,\; t_0+(K-1)M\},
\end{equation*}
where $t_0$ is chosen after the model has completed its initial burn-in phase as detailed in Algorithm~\ref{alg:dsi}.

If the trajectory evolves smoothly under a distributional metric, consecutive checkpoints induce highly correlated components. Increasing the stride $M$ increases the separation in parameter space between retained checkpoints and can increase the effective checkpoint count by lowering tail-error autocorrelation. As $K$ grows and the mixture spans a larger portion of the trajectory, the marginal impact of the stride naturally diminishes. We study this effect empirically in Section~\ref{sec:results}.

\subsection{The Bias-Variance Trajectory Averaging Hypothesis}
\label{subsec:tap}
The fundamental advantage of checkpoint ensembling is not that the mixture $\bar P_S$ is automatically unbiased. Instead, the mixture aggregates a sequence of checkpoint-specific tail errors. If these errors fluctuate around a persistent trajectory bias and decorrelate under checkpoint spacing, then averaging reduces the fluctuation component of tail-risk error. If the errors remain highly correlated or share the same directional misspecification, DSI cannot correct the failure.

This motivates the \emph{Bias-Variance Trajectory Averaging Principle}: under local tail regularity, finite-budget DSI error decomposes into (i) a persistent bias floor of the checkpoint trajectory, (ii) a checkpoint-fluctuation term controlled by an effective checkpoint count, and (iii) a synthetic-sampling term induced by the fixed generation budget. Section~\ref{sec:theory} formalizes this principle for Value-at-Risk and Expected Shortfall.

\section{Theoretical Analysis}
\label{sec:theory}

We now give a finite-budget bias-variance account of DSI. DSI targets the checkpoint-mixture distribution $\bar P_S$, not $P^\star$ directly. Its usefulness therefore rests on two conditions: the mixture $\bar P_S$ must remain close to the true target distribution $P^\star$, and checkpoint-specific tail errors must fluctuate sufficiently along the training trajectory. Under these conditions, pooling samples across spaced checkpoints reduces the fluctuation-driven component of tail-functional estimation error, while any persistent mixture bias relative to $P^\star$ remains as an irreducible floor.

\paragraph{Setup.}
Let $F^\star$ denote the CDF of the true scalar target distribution $P^\star$, and let $F_t$ denote the CDF induced by checkpoint $t$. For a selected checkpoint set $S$, DSI targets the checkpoint-mixture CDF $\bar F_S(x):=|S|^{-1}\sum_{t\in S}F_t(x)$, equivalently the mixture distribution $\bar P_S:=|S|^{-1}\sum_{t\in S}P_t$. The empirical DSI estimator draws a fixed total budget of synthetic samples across checkpoints in $S$ and applies the plug-in tail functional to the pooled empirical distribution $\widehat P_{S,N}$.

With $q^\star:=q_\alpha(P^\star)$ denoting the true lower-tail quantile, the VaR and ES checkpoint-level tail-error variables are
\begin{align}
    U_t^q &= F_t(q^\star)-F^\star(q^\star),\\
    U_t^{ES} &= \int_{-\infty}^{q^\star}\!\bigl(F_t(x)-F^\star(x)\bigr)\,dx.
\end{align}
Under the post-burn-in checkpoint-error assumption in Appendix~\ref{subsec:assumptions}, each target functional $\psi\in\{q_\alpha,\ES_\alpha\}$ has an associated checkpoint-error process $U_t^\psi=b_\psi+\xi_t^\psi$, where $b_\psi$ is the persistent bias and $\xi_t^\psi$ is a mean-zero weakly stationary fluctuation with variance $\sigma_\psi^2$ and autocorrelation $\rho_\psi(h)$. The effective checkpoint count for $\psi$, with $K_S=|S|$, is
\begin{equation}
    K_{\mathrm{eff},\psi}(K_S,M)
    =
    \frac{K_S}{\displaystyle 1+2\sum_{\ell=1}^{K_S-1}\!\left(1-\frac{\ell}{K_S}\right)\rho_\psi(\ell M)}.
\end{equation}
Thus $K_{\mathrm{eff},\psi}$ is the number of effectively independent checkpoint errors for the functional being estimated: it is close to $K_S$ when spaced checkpoint errors are weakly correlated and close to one when they are highly correlated. In the theorem below we apply this construction to $\psi=q_\alpha$ and $\psi=\ES_\alpha$, writing $K_{\mathrm{eff},q}$ and $K_{\mathrm{eff},ES}$ and suppressing their $(K_S,M)$ arguments. The empirical diagnostics in Appendix~\ref{sec:checkpoint_diagnostics} estimate these quantities for the trained trajectories.

The assumptions have three roles. Assumption~\ref{ass:tail_regular} gives local tail regularity, ensuring that small CDF perturbations near $q^\star$ translate smoothly into VaR and ES perturbations. Assumptions~\ref{ass:checkpoint_process} and~\ref{ass:two_stage_regularity} describe the post-burn-in checkpoint-error process and the local regularity of the selected checkpoint mixture, separating persistent trajectory bias from averageable checkpoint fluctuations. Assumptions~\ref{ass:sampling_moments} and~\ref{ass:stratified_bahadur} handle the finite-budget stratified sampling layer: DSI samples are independent conditional on checkpoints but not identically distributed marginally, and they estimate the mixture $\bar P_S$, not a single checkpoint. Full technical statements are given in Appendix~\ref{subsec:assumptions}.

\begin{theorem}[Finite-budget DSI MSE expansion]
\label{thm:finite_budget}
Let $\widehat q^{DSI}_{\alpha,S,N}=q_\alpha(\widehat P_{S,N})$ and $\widehat\ES^{DSI}_{\alpha,S,N}=\ES_\alpha(\widehat P_{S,N})$. Under the assumptions stated in Appendix~\ref{subsec:assumptions},
\begin{align}
\begin{split}
&\E\!\left[\bigl(\widehat q^{DSI}_{\alpha,S,N}-q^\star\bigr)^2\right] \\
&\quad = \frac{1}{f^\star(q^\star)^2} \!\left(b_q^2+\frac{\sigma_q^2}{K_{\mathrm{eff},q}}\right) +L^{sam}_{q,N}+o(\cdot),
\end{split} \\[6pt]
\begin{split}
&\E\!\left[\bigl(\widehat\ES^{DSI}_{\alpha,S,N}-\ES_\alpha(P^\star)\bigr)^2\right] \\
&\quad = \frac{1}{\alpha^2} \!\left(b_{ES}^2+\frac{\sigma_{ES}^2}{K_{\mathrm{eff},ES}}\right) +L^{sam}_{ES,N}+o(\cdot).
\end{split}
\end{align}
Here $L^{sam}_{q,N}$ and $L^{sam}_{ES,N}$ are the stratified finite-sample plug-in variances for the fixed checkpoint mixture. Their exact forms, the full two-stage theorem statement, and proof are given in Appendix~\ref{sec:proofs}.
\end{theorem}

\paragraph{Interpretation.}
The expansion separates DSI error into a trajectory term and a sampling term. For a tail functional $\psi$, the trajectory term $b_\psi^2+\sigma_\psi^2/K_{\mathrm{eff},\psi}$ decomposes into a persistent bias floor $b_\psi^2$ and a checkpoint-fluctuation component $\sigma_\psi^2/K_{\mathrm{eff},\psi}$ that decreases as $K_{\mathrm{eff},\psi}$ grows. DSI targets the checkpoint mixture, not $P^\star$ directly, so bias is not removed by averaging. Stride $M$ helps only by reducing autocorrelation, increasing $K$ helps only while it increases the functional-specific $K_{\mathrm{eff},\psi}$, and very large $K$ is eventually limited by floor allocation and the constraint $K_S\le N$.

Two immediate consequences make the theorem operational: it predicts when DSI should help and when trajectory averaging must fail.

\paragraph{Corollary 1 (Mechanistic regimes).}
Theorem~\ref{thm:finite_budget} identifies three regimes for each tail functional $\psi$: (i) a variance-reducible regime where $b_\psi^2$ is small and $K_{\mathrm{eff},\psi}$ grows, (ii) a correlation-limited regime where checkpoint fluctuations are present but highly autocorrelated, so $K_{\mathrm{eff},\psi}\ll K_S$, and (iii) a bias-limited regime where $b_\psi^2$ dominates and DSI cannot remove the leading error term. The proof is given in Appendix~\ref{subsec:proof_corollary}.

\paragraph{Proposition 1 (Failure modes).}
DSI cannot correct perfectly correlated checkpoint errors, shared directional lower-tail misspecification across all selected checkpoints, or persistent bias floors. In particular, if $\rho_\psi(\ell M)=1$ for all $\ell=1,\ldots,K_S-1$, then $K_{\mathrm{eff},\psi}=1$. If, for some sign $s\in\{-1,1\}$ and nonnegative function $d$, $s\{F_t(x)-F^\star(x)\}\ge d(x)$ for all $t\in S$ on a tail interval, then $s\{\bar F_S(x)-F^\star(x)\}\ge d(x)$ on that interval. Even when $K_{\mathrm{eff},\psi}\to\infty$ and $N\to\infty$, the first-order VaR and ES MSE floors remain $b_q^2/f^\star(q^\star)^2$ and $b_{ES}^2/\alpha^2$, respectively. The proof is given in Appendix~\ref{subsec:proof_failure_modes}.

These regimes motivate the diagnostics in Appendix~\ref{sec:checkpoint_diagnostics} and the sensitivity study in Figure~\ref{fig:sensitivity_summary}: DSI should help most when checkpoint errors for the target functional decorrelate across the trajectory, and should plateau when $K_{\mathrm{eff},\psi}$ saturates or the bias floor dominates, see also Appendix Figure~\ref{fig:autocorr_decay}.

\section{Experimental Setup}
\label{sec:experiments}

Our theoretical analysis decomposes finite-budget tail-risk error under the trajectory-induced mixture $\bar P_S$ into trajectory bias, checkpoint-fluctuation variance, and synthetic-sampling variance. We evaluated DSI empirically in this regime to assess: 
(i) stabilization of tail-functional estimation relative to single-checkpoint inference, 
(ii) preservation of structural properties (cross-asset and temporal dependence), and 
(iii) robustness to checkpoint selection parameters (mixture size $K$ and stride $M$).

Following the experimental protocol of \citet{Cont2025}, tail-dependent functionals are evaluated on profit-and-loss (PnL) distributions induced by $J$ benchmark trading strategies applied to multivariate return processes. The univariate functionals $q_\alpha$ and $\ES_\alpha$ analyzed in Section~\ref{sec:theory} are thus computed on strategy-level empirical distributions.

\subsection{Data and Fixed-Budget Regime}
\label{subsec:data_eval}

\paragraph{Synthetic data (oracle setting).}
Following \citet{Cont2025}, we construct a five-dimensional return process $\mathbf{X}(t)$ combining heterogeneous components (Gaussian, AR(1), and heavy-tailed Student-$t$ GARCH(1,1)) with fixed multivariate correlation. Because the true data-generating distribution $P^\star$ is analytically known, exact oracle values of $q_\alpha(P^\star)$ and $\ES_\alpha(P^\star)$ are available for direct error evaluation. Full parameter specifications for the synthetic process are detailed in Appendix~\ref{subsec:synthetic_setup}.

\paragraph{Market data (empirical setting).}
Following the experimental setup and asset selection criteria established in \citet{Cont2025}, we utilized the Nasdaq ITCH limit order book dataset for five liquid assets (AAPL, AMZN, GOOG, JPM, QQQ), sampled every $\Delta=9$ seconds. Each trajectory spans 100 time steps (15 minutes). Training used November 2019 data (6,300 samples), and testing uses the first week of December 2019 for out-of-sample evaluation. The empirical ground-truth risk measures (VaR and ES) for these assets across benchmark strategies are provided in Appendix Table~\ref{tab:market_data_summary}.

\paragraph{Fixed-budget regime.}
To isolate estimator variance and bias, all generative methods are evaluated under a fixed sampling budget of $N=1000$ generated trajectories, reflecting practical risk-management constraints. For DSI, drawing from the selected checkpoint set \(S\) implies \(m=\lfloor N/K_S\rfloor\) samples per checkpoint, directly matching the stratified finite-budget analysis in Theorem~\ref{thm:finite_budget}.

\subsection{Baselines and References}
\label{subsec:baselines}

We compared DSI against general adversarial generation (\textbf{WGAN} \citep{arjovsky2017wasserstein} with gradient penalty \citep{gulrajani2017improved}), explicitly tail-optimized generation (\textbf{TailGAN} \citep{Cont2025}), and standard single-checkpoint diffusion inference (\textbf{DDPM (EMA)} \citep{Ho2020}). 

The DDPM (EMA) baseline generates samples using an exponential moving average of the network weights, a common practice in diffusion-model training that smooths high-frequency optimization noise. We use the best EMA checkpoint as the single-checkpoint diffusion baseline, and apply DSI to the same training trajectory without changing the architecture or objective. Full implementation details are provided in Appendix~\ref{app:default_diffusion}.

Historical Simulation (\textbf{HSM}) is a nonparametric risk-estimation reference that computes VaR and ES directly from the empirical distribution of historical PnL values, without fitting a generative model. It serves as a conventional finance baseline rather than a learned simulator. On synthetic data, \textbf{SE(1000)} denotes the irreducible sampling error obtained by drawing exactly 1000 samples from the true synthetic process. On market data, \textbf{Oracle} denotes risk estimates computed from the full empirical reference set.

\subsection{Evaluation Metrics}
\label{subsec:metrics}
\paragraph{Benchmark Trading Strategies.}
Following the experimental protocol of \citet{Cont2025}, we evaluated downstream tail risk on profit-and-loss (PnL) distributions induced by a set of $J=65$ canonical trading strategies. These strategies instantiate the random variable $X$ analyzed in Section~\ref{sec:theory} by mapping multivariate return paths into realistic financial outcomes. We categorize these mappings into three functional groups: (i) \textbf{Buy-and-Hold} (testing heavy-tailed multivariate dependence), (ii) \textbf{Mean Reversion} (testing short-term negative autocorrelation), and (iii) \textbf{Trend Following} (testing long-term momentum). These strategies serve as deterministic operators applied identically to empirical reference data and synthetic trajectories, providing the empirical basis for computing Value-at-Risk ($q_\alpha$) and Expected Shortfall ($\ES_\alpha$). The precise mathematical formulations and dynamic trading rules are detailed in Appendix~\ref{subsec:strategies}.

\paragraph{Tail Accuracy.}
For confidence level $\alpha$, the average relative error (RE) of VaR and ES across all $J$ strategies is:

\begin{equation}
\begin{split}
\mathrm{RE}(\alpha) = \frac{1}{2J} \sum_{j=1}^{J} & \left( \frac{\big| \widehat{\VaR}_\alpha(\tilde{\mathcal{X}}_j) - \VaR_\alpha(\mathcal{X}_j) \big|}{\big| \VaR_\alpha(\mathcal{X}_j) \big|} \right. \\
& \left. + \frac{\big| \widehat{\ES}_\alpha(\tilde{\mathcal{X}}_j) - \ES_\alpha(\mathcal{X}_j) \big|}{\big| \ES_\alpha(\mathcal{X}_j) \big|} \right)
\end{split}
\end{equation}

where $\mathcal{X}_j$ and $\tilde{\mathcal{X}}_j$ denote the PnL distributions for strategy $j$ computed from reference and generated data, respectively.

\paragraph{Backtesting of Tail-Risk Estimates.}
We evaluate distributional consistency via statistical backtests standard in risk management, conducted on synthetic data over 100 trials using $N=1000$ generated samples. The Kupiec coverage test \citep{kupiec1995techniques} evaluates whether the realized frequency of VaR violations matches the nominal tail level $\alpha$. For example, a calibrated 5\% VaR should be breached approximately 5\% of the time. Since VaR alone does not measure the severity of losses beyond the threshold, we also use the Fissler-Ziegel score-based test \citep{Fissler2016}, which jointly evaluates VaR and ES calibration through a strictly consistent scoring function. We report rejection rates, where lower values indicate better agreement with the target tail-risk behavior.

\subsection{Implementation Details}
\label{subsec:implementation_details}
All reported results are averaged over 10 independent training runs with different random seeds. Unless stated otherwise, DSI uses the fixed budget ($N=1000$) with mixture size $K=20$. Additional details are provided in Appendix~\ref{sec:add_implementation}.

\section{Experimental Results}
\label{sec:results}
\paragraph{Accuracy and Stability of Tail-Risk Estimation.}

\begin{table}[t]
\centering
\small
\caption{Average Relative Error (RE) of VaR and ES ($\alpha=0.05$) under a fixed budget of $N=1000$ generated trajectories. SE(1000) denotes the irreducible sampling error on synthetic data. Lower is better.}
\label{tab:merged_re}
\begin{tabular}{lcc}
\toprule
Model & Synthetic RE & Market RE \\
\midrule
HSM & 3.4 $\pm$ 2.6 & 11.4 $\pm$ 2.3 \\
SE(1000) / Oracle & 3.1 $\pm$ 2.3 & 2.6 $\pm$ 1.1 \\
\midrule
WGAN & 17.3 $\pm$ 2.2 & 29.9 $\pm$ 1.9 \\
TailGAN & 5.1 $\pm$ 1.4 & 12.1 $\pm$ 1.3 \\
DDPM (EMA) & 7.7 $\pm$ 2.5 & 22.3 $\pm$ 4.3 \\ 
\midrule
WGAN (DSI) & 16.6 $\pm$ 1.8 & 28.5 $\pm$ 1.4 \\
TailGAN (DSI) & 5.0 $\pm$ 1.3 & 11.9 $\pm$ 1.2 \\
\textbf{DDPM (DSI)} & \textbf{3.7 $\pm$ 0.12} & \textbf{10.6 $\pm$ 1.2} \\
\bottomrule
\end{tabular}
\end{table}

Table~\ref{tab:merged_re} reports tail-estimation error under the fixed sampling budget. On synthetic data, standard single-checkpoint DDPM (EMA) exhibits substantial error and high variance ($7.7 \pm 2.5$). Checkpoint ensembling via DSI reduces the mean error to $3.7$ and sharply reduces variance ($\pm 0.12$), performing near the sampling-error reference SE(1000). On market data, DSI reduces single-checkpoint DDPM error ($10.8 \pm 1.2$ vs.\ $22.3 \pm 4.3$) and outperforms the explicitly tail-optimized TailGAN without modifying the underlying objective. Rank-frequency PnL plots in Appendix Figures~\ref{fig:synthetic_rankfreq} and \ref{fig:market_rankfreq} visually support the improved lower-tail fit. Additional sensitivity results across \(\alpha \in \{0.01,0.05,0.10\}\) are reported in Appendix Table~\ref{tab:alpha_sensitivity}, showing the same qualitative improvement pattern under the fixed-budget protocol.

\paragraph{Checkpoint-error regimes.}
The heterogeneous gains in Table~\ref{tab:merged_re} are consistent with the mechanistic regimes in Corollary~\ref{cor:regimes}, rather than indicating that DSI is tied to a particular architecture. DSI helps when the selected trajectory has a small persistent tail-bias floor and sufficiently decorrelated checkpoint fluctuations. The diagnostics in Appendix~\ref{sec:checkpoint_diagnostics} make this distinction explicit, see Appendix Figure~\ref{fig:autocorr_decay}. In Table~\ref{tab:es_diagnostics}, DDPM has a small bias floor, moderate checkpoint fluctuation variance, and a high effective checkpoint ratio \(K_{\mathrm{eff}}/K=0.953\), placing it in the variance-reducible regime. TailGAN has substantial fluctuation variance, but \(K_{\mathrm{eff}}/K=0.149\), indicating a correlation-limited trajectory where nearby checkpoint errors are too redundant for pooling to provide large gains. WGAN has high sign consensus, indicating shared error direction across checkpoints, and its finite-budget error in Table~\ref{tab:finite_budget_diagnostics} is dominated by the mixture-error MSE rather than the sampling term, indicating a bias-limited or shared-direction-error regime. These observations support Proposition~\ref{prop:failure_modes}: the relevant condition for DSI is not the model family itself, but whether the checkpoint trajectory contains averageable tail variation.

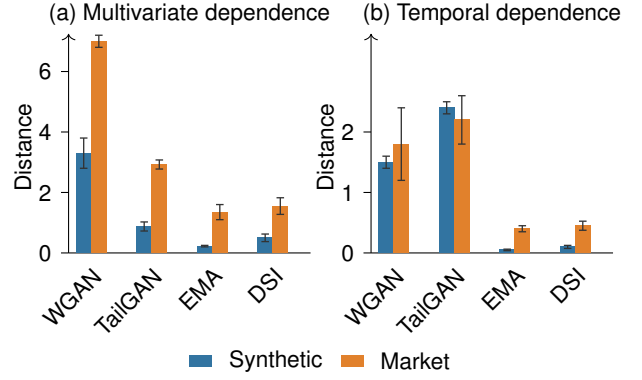
\begin{figure}[t]
    \centering

\begin{tikzpicture}[
    font=\sffamily\small,           
    small/.style={font=\sffamily\small}, 
    legend/.style={font=\sffamily\footnotesize}, 
    errorbar/.style={line width=0.5pt, draw=black!80}
]

\def\W{3.2}    
\def\H{2.8}    
\def\G{0.8}    

\definecolor{colorSyn}{HTML}{3274A1} 
\definecolor{colorMkt}{HTML}{E1812C} 

\newcommand{\errbar}[4]{
    \draw[errorbar] (#1,#2) -- (#1,#3);
    \draw[errorbar] (#1-#4,#2) -- (#1+#4,#2);
    \draw[errorbar] (#1-#4,#3) -- (#1+#4,#3);
}

\begin{scope}[shift={(0,0)}]
    \node[small, anchor=south] at (\W/2, \H+0.1) {(a) Multivariate dependence};
    \draw[->] (0,0) -- (0,\H+0.1);
    \node[small, rotate=90, anchor=center] at (-0.6, {0.5*\H}) {Distance};
    \draw (0,0) -- (\W,0);

    \foreach \y/\label in {0/0, 0.8/2, 1.6/4, 2.4/6}
        \draw (0,\y) -- (-0.1,\y) node[small, left] {\label};

    \fill[colorSyn] (0.1,0) rectangle (0.3, 1.32); 
    \fill[colorMkt] (0.3,0) rectangle (0.5, 2.80);
    \errbar{0.2}{1.12}{1.52}{0.05}
    \errbar{0.4}{2.72}{2.88}{0.05}
    
    \fill[colorSyn] (0.9,0) rectangle (1.1, 0.35); 
    \fill[colorMkt] (1.1,0) rectangle (1.3, 1.17);
    \errbar{1.0}{0.29}{0.41}{0.05}
    \errbar{1.2}{1.11}{1.23}{0.05}

    \fill[colorSyn] (1.7,0) rectangle (1.9, 0.09); 
    \fill[colorMkt] (1.9,0) rectangle (2.1, 0.54);
    \errbar{1.8}{0.08}{0.10}{0.05}
    \errbar{2.0}{0.44}{0.64}{0.05}

    \fill[colorSyn] (2.5,0) rectangle (2.7, 0.20); 
    \fill[colorMkt] (2.7,0) rectangle (2.9, 0.62);
    \errbar{2.6}{0.15}{0.25}{0.05}
    \errbar{2.8}{0.51}{0.73}{0.05}

    \node[legend, rotate=45, anchor=north east] at (0.3,0) {WGAN};
    \node[legend, rotate=45, anchor=north east] at (1.1,0) {TailGAN};
    \node[legend, rotate=45, anchor=north east] at (1.9,0) {EMA};
    \node[legend, rotate=45, anchor=north east] at (2.7,0) {DSI};
\end{scope}

\begin{scope}[shift={(\W+\G,0)}]
    \node[small, anchor=south] at (\W/2, \H+0.1) {(b) Temporal dependence};
    \draw[->] (0,0) -- (0,\H+0.1);
    \node[small, rotate=90, anchor=center] at (-0.6, {0.5*\H}) {Distance};
    \draw (0,0) -- (\W,0);

    \foreach \y/\label in {0/0, 0.8/1, 1.6/2}
        \draw (0,\y) -- (-0.1,\y) node[small, left] {\label};

    \fill[colorSyn] (0.1,0) rectangle (0.3, 1.20); 
    \fill[colorMkt] (0.3,0) rectangle (0.5, 1.44);
    \errbar{0.2}{1.12}{1.28}{0.05}
    \errbar{0.4}{0.96}{1.92}{0.05}
    
    \fill[colorSyn] (0.9,0) rectangle (1.1, 1.92); 
    \fill[colorMkt] (1.1,0) rectangle (1.3, 1.76);
    \errbar{1.0}{1.84}{2.00}{0.05}
    \errbar{1.2}{1.44}{2.08}{0.05}

    \fill[colorSyn] (1.7,0) rectangle (1.9, 0.04); 
    \fill[colorMkt] (1.9,0) rectangle (2.1, 0.32);
    \errbar{1.8}{0.03}{0.05}{0.05}
    \errbar{2.0}{0.28}{0.36}{0.05}

    \fill[colorSyn] (2.5,0) rectangle (2.7, 0.08); 
    \fill[colorMkt] (2.7,0) rectangle (2.9, 0.36);
    \errbar{2.6}{0.06}{0.10}{0.05}
    \errbar{2.8}{0.30}{0.42}{0.05}

    \node[legend, rotate=45, anchor=north east] at (0.3,0) {WGAN};
    \node[legend, rotate=45, anchor=north east] at (1.1,0) {TailGAN};
    \node[legend, rotate=45, anchor=north east] at (1.9,0) {EMA};
    \node[legend, rotate=45, anchor=north east] at (2.7,0) {DSI};
\end{scope}

\begin{scope}[shift={(\W/2, -1.5)}]
    \fill[colorSyn] (0,0) rectangle (0.3, 0.15);
    \node[legend, anchor=west] at (0.4, 0.07) {Synthetic};
    \fill[colorMkt] (2.0,0) rectangle (2.3, 0.15);
    \node[legend, anchor=west] at (2.4, 0.07) {Market};
\end{scope}
\end{tikzpicture}
    \caption{Dependence preservation under realistic sampling budgets.
    (a) Sum of absolute differences between reference and generated correlation matrices (multivariate dependence).
    (b) Sum of absolute differences between reference and generated autocorrelation coefficients (temporal dependence).
    Lower values indicate better agreement.
    Error bars denote standard deviation across runs.}
    \label{fig:dependence_summary}
\end{figure}

\paragraph{Backtesting and Structural Fidelity.}
\label{para:results_backtests}

\begin{table}[t]
\centering
\small
\caption{Average rejection rate (\%) for statistical backtests over 100 trials ($N=1000$). Lower rejection rates indicate superior calibration.}
\label{tab:stat_tests}
\begin{tabular}{lcc}
\toprule
Model & Coverage Test & Score-Based Test \\
\midrule
HSM & 17.19 & 0.00 \\
Oracle & 16.32 & 0.00 \\
\midrule
WGAN & 70.35 & 14.92 \\
TailGAN & 33.55 & 0.29 \\
DDPM (EMA) & 53.62 & 0.06 \\
\textbf{DDPM (DSI)} & \textbf{20.26} & \textbf{0.00} \\
\bottomrule
\end{tabular}
\vspace{-2mm} 
\end{table}

Table~\ref{tab:stat_tests} reports formal risk backtests. Standard diffusion fails the coverage test in 53.62\% of trials, reflecting lower-tail miscalibration. DSI reduces this rejection rate to 20.26\% (approaching the Oracle's 16.32\%) and achieves 0.00\% rejection on the joint VaR/ES Fissler-Ziegel test, indicating improved probabilistic calibration under the fixed budget.

\paragraph{Multivariate and Temporal Dependency.}
DSI also preserves multivariate and temporal dependence metrics. Analyzing absolute differences between generated and reference correlation matrices and temporal autocorrelations (Figure~\ref{fig:dependence_summary}), DSI retains the structural fidelity of the underlying diffusion model and outperforms GAN-based methods on these diagnostics. This suggests that sample-space ensembling can improve marginal tail estimates without visibly degrading learned multivariate geometries. Full results for correlation and autocorrelation matrices are provided in Appendix Figures~\ref{fig:synthetic_correlation}-\ref{fig:market_autocorr}.

\paragraph{Sensitivity to Mixture Size and Stride.}
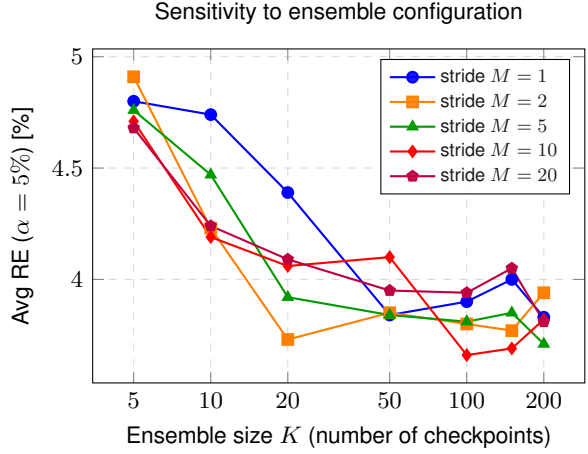
\begin{figure}[t]
    \centering

\begin{tikzpicture}[
    font=\sffamily\small,
    small/.style={font=\sffamily\small},
    legend/.style={font=\sffamily\tiny},
]

\begin{axis}[
    width=0.98\columnwidth,
    height=6cm,
    title={Sensitivity to ensemble configuration},
    xlabel={Ensemble size $K$ (number of checkpoints)},
    ylabel={Avg RE ($\alpha=5\%$) [\%]},
    xmode=log,
    log ticks with fixed point,
    xtick={5, 10, 20, 50, 100, 200},
    grid=major,
    grid style={dashed, gray!30},
    legend pos=north east,
    legend cell align={left},
    legend style={nodes={scale=0.8}, fill opacity=0.9}
]
    \addplot[blue, mark=*, thick] coordinates {
        (5,4.80) (10,4.74) (20,4.39) (50,3.84) (100,3.90) (150,4.00) (200,3.83)
    }; \addlegendentry{stride $M=1$}

    \addplot[orange, mark=square*, thick] coordinates {
        (5,4.91) (10,4.23) (20,3.73) (50,3.85) (100,3.80) (150,3.77) (200,3.94)
    }; \addlegendentry{stride $M=2$}

    \addplot[green!60!black, mark=triangle*, thick] coordinates {
        (5,4.76) (10,4.47) (20,3.92) (50,3.84) (100,3.81) (150,3.85) (200,3.71)
    }; \addlegendentry{stride $M=5$}

    \addplot[red, mark=diamond*, thick] coordinates {
        (5,4.71) (10,4.19) (20,4.06) (50,4.10) (100,3.66) (150,3.69) (200,3.82)
    }; \addlegendentry{stride $M=10$}

    \addplot[purple, mark=pentagon*, thick] coordinates {
        (5,4.68) (10,4.24) (20,4.09) (50,3.95) (100,3.94) (150,4.05) (200,3.81)
    }; \addlegendentry{stride $M=20$}

\end{axis}
\end{tikzpicture}
    \caption{Sensitivity of DSI to mixture size $K$ and checkpoint stride $M$ under a fixed budget ($N=1000$).
    Average relative error (VaR and ES, $\alpha=5\%$) is plotted against $K$ for different strides.
    Checkpoint values on the x-axis are displayed with uniform spacing.
    Error decreases rapidly with increasing $K$ and plateaus for large mixtures.
    Larger strides provide gains when $K$ is small by reducing checkpoint correlation, with diminishing impact as $K$ grows.
    }
    \label{fig:sensitivity_summary}
\end{figure}

Finally, we empirically evaluate DSI's sensitivity to the checkpoint selection parameters introduced in Section~\ref{subsec:ckp_stride}, the mixture size $K$ and sampling stride $M$. As shown in Figure~\ref{fig:sensitivity_summary} and Appendix Table~\ref{tab:full_sensitivity_results}, risk-estimation error improves rapidly as $K$ increases from 5 to roughly 50, after which performance plateaus under the fixed sampling budget. This behavior is consistent with Theorem~\ref{thm:finite_budget}: increasing $K$ helps while it increases the effective checkpoint count, but gains plateau once $K_{\mathrm{eff}}$ saturates and finite-budget remainders become non-negligible. Larger strides improve performance when $K$ is small by reducing checkpoint-error autocorrelation, see Appendix Figure~\ref{fig:autocorr_decay}. As $K$ grows and the mixture spans a larger portion of the trajectory, the marginal effect of stride diminishes.

\section{Discussion and Conclusion}
\label{sec:discussion_conclusion}


We introduced Diachronic Sample Integration (DSI), a lightweight test-time estimator for finite-budget tail-risk estimation with generative simulators. By reusing stored checkpoints from a single stochastic training trajectory, DSI pools generated samples to target a trajectory-induced mixture distribution. Our theory and diagnostics show that DSI reduces checkpoint-fluctuation error when the trajectory contains averageable tail variation, while persistent bias or strong autocorrelation remains irreducible. Empirically, DSI improves risk estimation error under fixed simulation budgets and achieves
state-of-the-art calibration without changing the generative objective.

\textbf{Scope of DSI.}
DSI is not tied to a particular generative architecture, its effectiveness is governed by whether the post-burn-in checkpoint trajectory contains averageable tail variation. As formalized in Corollary~\ref{cor:regimes} and Proposition~\ref{prop:failure_modes}, DSI helps in variance-reducible regimes but remains limited by persistent bias or strong checkpoint-error autocorrelation.

\textbf{Broader impact.} While our empirical evaluation rigorously focused on high-frequency financial time series, a canonical stress test due to their extreme signal-to-noise ratios, heavy tails, and complex multivariate dependencies, the mathematical formulation of DSI is entirely domain-agnostic. Theory derived in Section~\ref{sec:theory} apply to any setting where exact sampling is infeasible and downstream functionals depend on the extremes of a generative simulator. We anticipate DSI will be highly effective in other critical domains requiring robust extreme-value estimation under finite budgets, such as energy, climate modeling, structural engineering, and epidemiology.

\textbf{Limitations and future work.}
Several limitations point to natural future work. First, DSI is primarily a variance-reduction mechanism over checkpoint fluctuations, it does not directly correct systematic tail bias. When all selected checkpoints share the same missing tail support or directional misspecification, the bias floor remains. Combining trajectory pooling with tail-guided or tail-constrained training methods is a promising direction for reducing this systematic component. Second, although our diagnostics characterize when DSI should work, they are currently evaluated after generating checkpoint samples. A useful next step is to develop lightweight compatibility diagnostics that can predict, early in training or with a small validation budget, whether a given model trajectory is variance-reducible, correlation-limited, or bias-limited. Third, our current estimator uses simple equal-weighted checkpoint mixtures, validation-weighted mixtures, adaptive stride selection, and bias-aware checkpoint selection may further improve finite-budget performance.


\bibliography{uai2026-template}

\newpage

\onecolumn

\title{Diachronic Sample Integration: Robust Tail-Risk Estimation with Generative Models\\(Supplementary Material)}
\maketitle

\appendix

\section{Proofs}
\label{sec:proofs}
This appendix provides the full assumptions, formal theorem, and proofs for the finite-budget theory in Section~\ref{sec:theory}. The proof uses a two-stage decomposition: first compare the checkpoint mixture $\bar P_S$ with the truth $P^\star$, and then compare the empirical DSI estimator with the fixed mixture $\bar P_S$. This avoids assuming a direct empirical plug-in linearization around $P^\star$.

\subsection{Assumptions and Full Theorem Statement}
\label{subsec:assumptions}

\begin{assumption}[Local tail regularity]
\label{ass:tail_regular}
$F^\star$ has a unique $\alpha$-quantile $q^\star$ and is continuously differentiable on $I_\delta=[q^\star-\delta,q^\star+\delta]$, with density $f^\star(q^\star)>0$ and $\inf_{x\in I_\delta}f^\star(x)\ge c>0$. When second-order remainders are invoked, $f^\star$ is locally Lipschitz on $I_\delta$.
\end{assumption}

\begin{assumption}[Post-burn-in checkpoint-error process]
\label{ass:checkpoint_process}
Let $X_t\sim P_t$ and $X^\star\sim P^\star$. Define $e_t(x)=F_t(x)-F^\star(x)$,
\begin{align}
    U_t^q &= e_t(q^\star), \\
    U_t^{ES} &= \int_{-\infty}^{q^\star}e_t(x)\,dx
    = \E[(q^\star-X_t)_+]-\E[(q^\star-X^\star)_+].
\end{align}
After burn-in, assume $U_t^q=b_q+\xi_t^q$ and $U_t^{ES}=b_{ES}+\xi_t^{ES}$, where $b_q,b_{ES}$ are persistent bias terms and $\xi_t^q,\xi_t^{ES}$ are mean-zero weakly stationary fluctuations with variances $\sigma_q^2,\sigma_{ES}^2$ and autocorrelations $\rho_q(h),\rho_{ES}(h)$.
\end{assumption}

\begin{assumption}[Sampling and lower-tail moments]
\label{ass:sampling_moments}
Assume $K_S\le N$ and set $m=\lfloor N/K_S\rfloor$ and $N_S=K_Sm$. Conditional on the selected checkpoint distributions $\{P_t:t\in S\}$, the generated samples $\{X_{t,i}:t\in S,1\le i\le m\}$ are independent with $X_{t,i}\sim P_t$. For all quantiles appearing below, the lower-tail variables $(q-X_t)_+$ have uniformly bounded second moments, and the analogous condition holds under $P^\star$.
\end{assumption}

\begin{assumption}[Local mixture regularity]
\label{ass:two_stage_regularity}
Let
\begin{align}
    q_S &:= q_\alpha(\bar P_S), \\
    \bar e_S(x) &:= \bar F_S(x)-F^\star(x), \\
    \eta_S &:= \sup_{x\in I_\delta}|\bar e_S(x)|.
\end{align}
Define the two leading checkpoint-mixture errors
\begin{align}
    \bar U_{q,S} &:= \bar F_S(q^\star)-F^\star(q^\star), \\
    \bar U_{ES,S} &:= \int_{-\infty}^{q^\star}(\bar F_S(x)-F^\star(x))\,dx,
\end{align}
and the local oscillation of the mixture error around $q^\star$,
\begin{equation}
    \omega_S(r):=\sup_{\substack{x\in I_\delta\\ |x-q^\star|\le r}}
    |\bar e_S(x)-\bar e_S(q^\star)|.
\end{equation}
Along the selected post-burn-in windows, the mixture remains in a local perturbation neighbourhood of $P^\star$: $q_S\in I_\delta$, $\bar F_S(q_S)=\alpha$, and $\bar F_S$ has a density $\bar f_S$ in a neighbourhood of $q_S$ with $\bar f_S(q_S)\ge c_0>0$. The local oscillation and second-order terms are negligible relative to the leading checkpoint-mixture errors:
\begin{equation}
    \E\!\left[(\omega_S(\eta_S/c)+\eta_S^2)^2\right]
    +\E\!\left[|\bar U_{q,S}|(\omega_S(\eta_S/c)+\eta_S^2)\right]  = o\!\left(\E[\bar U_{q,S}^2]\right),
\end{equation}
and
\begin{equation}
    \E[\eta_S^4]+\E[|\bar U_{ES,S}|\eta_S^2]
    =o\!\left(\E[\bar U_{ES,S}^2]\right).
\end{equation}
Assumption~\ref{ass:two_stage_regularity} is a local perturbation condition: it requires the selected post-burn-in mixture to remain in a neighbourhood where VaR and ES admit stable first-order expansions.
\end{assumption}

\begin{assumption}[Standard stratified plug-in linearization]
\label{ass:stratified_bahadur}
Conditional on the selected checkpoint distributions, the empirical DSI estimators satisfy the usual Bahadur/plug-in linearizations around the fixed mixture $\bar P_S$:
\begin{equation}
\widehat q^{DSI}_{\alpha,S,N}-q_S
= -\frac{\widehat F_{S,N}(q_S)-\alpha}{\bar f_S(q_S)}+r^{sam}_{q,S,N},
\end{equation}
\begin{equation}
\widehat\ES^{DSI}_{\alpha,S,N}-\ES_\alpha(\bar P_S)
= -\frac{1}{\alpha}\int_{-\infty}^{q_S}(\widehat F_{S,N}(x)-\bar F_S(x))\,dx
+r^{sam}_{ES,S,N},
\end{equation}
with $\E[(r^{sam}_{q,S,N})^2\mid\{P_t\}]=o(N_S^{-1})$ and $\E[(r^{sam}_{ES,S,N})^2\mid\{P_t\}]=o(N_S^{-1})$ along the finite-budget sequences considered. These conditions hold under standard local density and moment assumptions for independent, stratified triangular arrays with positive density at the target mixture quantile. They are stated explicitly to separate sampling regularity from trajectory regularity in Assumption~\ref{ass:two_stage_regularity}.
\end{assumption}

\paragraph{Full theorem statement.}
For an autocorrelation function $\rho$, define
\begin{align}
    \kappa_\rho(K_S,M) &= 1+2\sum_{\ell=1}^{K_S-1}\left(1-\frac{\ell}{K_S}\right)\rho(\ell M), \\
    K_{\mathrm{eff},\rho}(K_S,M) &= \frac{K_S}{\kappa_\rho(K_S,M)},
\end{align}
assuming $\kappa_\rho(K_S,M)>0$. Define the conditional sampling variances at the mixture quantile
\begin{equation}
    A^{\mathrm{mix}}_{q,N}(S)=\frac{1}{K_S^2m}\sum_{t\in S}F_t(q_S)(1-F_t(q_S)),
\end{equation}
\begin{equation}
    A^{\mathrm{mix}}_{ES,N}(S)=\frac{1}{K_S^2m}\sum_{t\in S}\Var_{P_t}[(q_S-X_t)_+].
\end{equation}
For comparison, write $A_{q,N}(S)$ and $A_{ES,N}(S)$ for the same two quantities with $q_S$ replaced by $q^\star$. Unless explicitly conditioned, expectations below are taken over both the post-burn-in checkpoint-error process and the synthetic samples drawn conditionally from the selected checkpoints.

\setcounter{theorem}{0}
\begin{theorem}[Two-stage finite-budget DSI MSE expansion, full statement]
\label{thm:finite_budget_full}
Suppose Assumptions~\ref{ass:tail_regular}-\ref{ass:stratified_bahadur} hold. Define
\begin{align}
L_q^{traj} &= b_q^2+\frac{\sigma_q^2}{K_{\mathrm{eff},q}(K_S,M)}, &
L_{q,N}^{sam} &= \E\!\left[\frac{A^{\mathrm{mix}}_{q,N}(S)}{\bar f_S(q_S)^2}\right],
\end{align}
\begin{align}
L_{ES}^{traj} &= b_{ES}^2+\frac{\sigma_{ES}^2}{K_{\mathrm{eff},ES}(K_S,M)}, &
L_{ES,N}^{sam} &= \E A^{\mathrm{mix}}_{ES,N}(S).
\end{align}
Then
\begin{align}
    \E\!\left[\left(\widehat q^{DSI}_{\alpha,S,N}-q^\star\right)^2\right] &= \frac{L_q^{traj}}{f^\star(q^\star)^2}+L_{q,N}^{sam}+\mathcal R_{q,N},\\
    \E\!\left[\left(\widehat\ES^{DSI}_{\alpha,S,N}-\ES_\alpha(P^\star)\right)^2\right] &= \frac{(L_{ES}^{traj}+L_{ES,N}^{sam})}{\alpha^2}+\mathcal R_{ES,N},
\end{align}
where
\begin{align}
    \mathcal R_{q,N} &= o\!\left(\frac{L_q^{traj}}{f^\star(q^\star)^2}\right)+o(N_S^{-1}), \\
    \mathcal R_{ES,N} &= o\!\left(\frac{L_{ES}^{traj}}{\alpha^2}\right)+o(N_S^{-1}).
\end{align}
If, additionally, $\bar f_S(q_S)=f^\star(q^\star)+o_{\mathbb P}(1)$ and the stratified variances at $q_S$ are first-order equivalent to those at $q^\star$, this reduces to the simpler common-denominator form with $A_{q,N}(S)$ and $A_{ES,N}(S)$ evaluated at $q^\star$. In the notation of Theorem~\ref{thm:finite_budget}, $L^{sam}_{q,N}=L_{q,N}^{sam}$ and $L^{sam}_{ES,N}=L_{ES,N}^{sam}/\alpha^2$.
\end{theorem}

\begin{corollary}[Mechanistic regimes]
\label{cor:regimes}
Theorem~\ref{thm:finite_budget_full} identifies three regimes: (i) a \emph{variance-reducible} regime where the corresponding bias term is small and $K_{\mathrm{eff}}$ grows, (ii) a \emph{correlation-limited} regime where checkpoint fluctuations are present but highly autocorrelated, so $K_{\mathrm{eff}}\ll K_S$, and (iii) a \emph{bias-limited} regime where $b^2$ dominates and DSI cannot remove the leading error term.
\end{corollary}

\begin{proposition}[Failure modes]
\label{prop:failure_modes}
DSI cannot correct perfectly correlated checkpoint errors, shared directional lower-tail misspecification across all selected checkpoints, or persistent bias floors. In particular, if $\rho(\ell M)=1$ for all $\ell=1,\ldots,K_S-1$, then $K_{\mathrm{eff}}=1$. If, for some sign $s\in\{-1,1\}$ and nonnegative function $a$, $s\{F_t(x)-F^\star(x)\}\ge a(x)$ for all $t\in S$ on a tail interval, then $s\{\bar F_S(x)-F^\star(x)\}\ge a(x)$ on that interval. Even when $K_{\mathrm{eff}}\to\infty$ and $N\to\infty$, the first-order VaR and ES MSE floors remain $b_q^2/f^\star(q^\star)^2$ and $b_{ES}^2/\alpha^2$, respectively.
\end{proposition}

\subsection{Auxiliary Statements}

\begin{lemma}[Smooth first-order tail-functional expansions]
\label{lem:functional_expansions}
Let $G$ be a continuous CDF with finite lower-tail first moment, $q_G=q_\alpha(G)\in I_\delta$, and $G(q_G)=\alpha$. Write $g=G-F^\star$, $\Delta=q_G-q^\star$, $\eta=\sup_{x\in I_\delta}|g(x)|$, and
\begin{equation}
    \omega_G(r):=\sup_{\substack{x\in I_\delta\\ |x-q^\star|\le r}}|g(x)-g(q^\star)|.
\end{equation}
Under Assumption~\ref{ass:tail_regular}, $|\Delta|\le \eta/c$ and
\begin{equation}
    q_\alpha(G)-q^\star
    =
    -\frac{g(q^\star)}{f^\star(q^\star)}+O\!\left(\omega_G(\eta/c)+\eta^2\right),
\end{equation}
while
\begin{equation}
    \ES_\alpha(G)-\ES_\alpha(P^\star)
    =
    -\frac{1}{\alpha}\int_{-\infty}^{q^\star}g(x)\,dx+O(\eta^2).
\end{equation}
\end{lemma}

\begin{proposition}[Checkpoint-error averaging]
\label{prop:checkpoint_averaging}
Under Assumption~\ref{ass:checkpoint_process},
\begin{equation}
    \E\left[\left(\frac{1}{K_S}\sum_{t\in S}U_t^q\right)^2\right]
    =
    b_q^2+\frac{\sigma_q^2}{K_{\mathrm{eff},q}(K_S,M)},
\end{equation}
and
\begin{equation}
    \E\left[\left(\frac{1}{K_S}\sum_{t\in S}U_t^{ES}\right)^2\right]
    =
    b_{ES}^2+\frac{\sigma_{ES}^2}{K_{\mathrm{eff},ES}(K_S,M)}.
\end{equation}
\end{proposition}

\begin{proposition}[Stratified DSI sampling identities]
\label{prop:stratified_cdf}
Under Assumption~\ref{ass:sampling_moments}, condition on the selected checkpoint distributions. For any fixed $x\in\R$,
\begin{equation}
    \E[\widehat F_{S,N}(x)\mid\{P_t:t\in S\}]=\bar F_S(x),
\end{equation}
and
\begin{equation}
    \Var[\widehat F_{S,N}(x)\mid\{P_t:t\in S\}]
    =
    \frac{1}{K_S^2m}\sum_{t\in S}F_t(x)(1-F_t(x)).
\end{equation}
If $m=N/K_S$ exactly, then this variance is at most $\bar F_S(x)(1-\bar F_S(x))/N$. Moreover, for any fixed threshold $q$ with finite lower-tail second moments,
\begin{equation}
    Z_{ES,N}(q):=\int_{-\infty}^{q}(\widehat F_{S,N}(x)-\bar F_S(x))\,dx
\end{equation}
satisfies
\begin{equation}
    \E[Z_{ES,N}(q)\mid\{P_t:t\in S\}]=0,
\end{equation}
and
\begin{equation}
    \Var[Z_{ES,N}(q)\mid\{P_t:t\in S\}]
    =\frac{1}{K_S^2m}\sum_{t\in S}\Var_{P_t}[(q-X_t)_+].
\end{equation}
\end{proposition}

\subsection{Proof of Lemma~\ref{lem:functional_expansions}}

Because $G(q_G)=\alpha$ and $F^\star(q^\star)=\alpha$,
\begin{equation}
0=G(q_G)-F^\star(q^\star)
=F^\star(q^\star+\Delta)-F^\star(q^\star)+g(q_G).
\end{equation}
The lower density bound in Assumption~\ref{ass:tail_regular} implies
\begin{equation}
    c|\Delta|\le |F^\star(q^\star+\Delta)-F^\star(q^\star)|=|g(q_G)|\le \eta,
\end{equation}
so $|\Delta|\le \eta/c$. A Taylor expansion of $F^\star$ around $q^\star$ gives
\begin{equation}
0=f^\star(q^\star)\Delta+g(q^\star)+\{g(q_G)-g(q^\star)\}+O(\Delta^2).
\end{equation}
Since $|\Delta|\le\eta/c$, the oscillation term is bounded by $\omega_G(\eta/c)$ and $\Delta^2=O(\eta^2)$. Hence
\begin{equation}
\Delta=-\frac{g(q^\star)}{f^\star(q^\star)}+O\!\left(\omega_G(\eta/c)+\eta^2\right),
\end{equation}
which proves the VaR expansion.

For ES, use the quantile-integral/lower-tail CDF identity, valid for continuous distributions with finite lower-tail first moment,
\begin{align}
    \ES_\alpha(G) &= q_G-\frac{1}{\alpha}\int_{-\infty}^{q_G}G(x)\,dx, \\
    \ES_\alpha(P^\star) &= q^\star-\frac{1}{\alpha}\int_{-\infty}^{q^\star}F^\star(x)\,dx.
\end{align}
Subtracting gives
\begin{align}
\ES_\alpha(G)-\ES_\alpha(P^\star)
&=\Delta-\frac{1}{\alpha}\left[\int_{-\infty}^{q^\star}g(x)\,dx+\int_{q^\star}^{q^\star+\Delta}G(x)\,dx\right].
\end{align}
The second integral satisfies
\begin{equation}
\int_{q^\star}^{q^\star+\Delta}G(x)\,dx
=\alpha\Delta+O(\Delta^2)+O(\eta|\Delta|),
\end{equation}
where the last term bounds the perturbation contribution on $I_\delta$. Since $|\Delta|\le\eta/c$, the $\Delta$ terms cancel and the remainder is $O(\eta^2)$:
\begin{equation}
\ES_\alpha(G)-\ES_\alpha(P^\star)
=
-\frac{1}{\alpha}\int_{-\infty}^{q^\star}g(x)\,dx+O(\eta^2).
\end{equation}

\subsection{Proof of Proposition~\ref{prop:checkpoint_averaging}}

We prove the VaR statement and the ES statement is identical. By Assumption~\ref{ass:checkpoint_process},
\begin{align}
    U_t^q &= b_q+\xi_t^q, &
    \E[\xi_t^q] &= 0.
\end{align}
Thus
\begin{equation}
    \frac{1}{K_S}\sum_{t\in S}U_t^q
    =
    b_q+\frac{1}{K_S}\sum_{t\in S}\xi_t^q.
\end{equation}
Squaring and taking expectations gives
\begin{equation}
    \E\left[\left(\frac{1}{K_S}\sum_{t\in S}U_t^q\right)^2\right]
    =
    b_q^2+
    \E\left[\left(\frac{1}{K_S}\sum_{t\in S}\xi_t^q\right)^2\right],
\end{equation}
because the cross term vanishes. Writing $S=\{t_0+rM:r=0,\ldots,K_S-1\}$, weak stationarity gives
\begin{equation}
    \Cov(\xi_{t_0+rM}^q,\xi_{t_0+sM}^q)
    =
    \sigma_q^2\rho_q(|r-s|M).
\end{equation}
Therefore,
\begin{align}
    \E\left[
    \left(\frac{1}{K_S}\sum_{r=0}^{K_S-1}\xi_{t_0+rM}^q\right)^2
    \right]
    & =
    \frac{\sigma_q^2}{K_S^2}
    \sum_{r=0}^{K_S-1}\sum_{s=0}^{K_S-1}\rho_q(|r-s|M) \nonumber\\
    & =
    \frac{\sigma_q^2}{K_S}
    \left[
        1+2\sum_{\ell=1}^{K_S-1}
        \left(1-\frac{\ell}{K_S}\right)\rho_q(\ell M)
    \right] \nonumber\\
    & =
    \frac{\sigma_q^2}{K_{\mathrm{eff},q}(K_S,M)}.
\end{align}
This proves the VaR identity. Replacing $U_t^q$ by $U_t^{ES}$ proves the ES identity.

\subsection{Proof of Proposition~\ref{prop:stratified_cdf}}

Condition on the selected checkpoint distributions. For any fixed $x$,
\begin{align}
    \widehat F_{S,N}(x) &= \frac{1}{K_S}\sum_{t\in S}\widehat F_{t,m}(x), \\
    \widehat F_{t,m}(x) &= \frac{1}{m}\sum_{i=1}^m\1\{X_{t,i}\le x\}.
\end{align}
Because $\E[\widehat F_{t,m}(x)\mid P_t]=F_t(x)$,
\begin{equation}
    \E[\widehat F_{S,N}(x)\mid\{P_t:t\in S\}]
    =\frac{1}{K_S}\sum_{t\in S}F_t(x)=\bar F_S(x).
\end{equation}
Conditional independence across samples and checkpoints gives
\begin{equation}
    \Var[\widehat F_{S,N}(x)\mid\{P_t:t\in S\}]
    =\frac{1}{K_S^2}\sum_{t\in S}\Var[\widehat F_{t,m}(x)\mid P_t]
    =\frac{1}{K_S^2m}\sum_{t\in S}F_t(x)(1-F_t(x)).
\end{equation}
If $m=N/K_S$, this becomes
\begin{equation}
    \frac{1}{K_SN}\sum_{t\in S}F_t(x)(1-F_t(x)).
\end{equation}
Since $p\mapsto p(1-p)$ is concave on $[0,1]$, Jensen's inequality yields
\begin{equation}
    \frac{1}{K_S}\sum_{t\in S}F_t(x)(1-F_t(x))
    \le \bar F_S(x)(1-\bar F_S(x)).
\end{equation}

For the integrated identity, fix $q$. For any sample value $X$,
\begin{equation}
    \int_{-\infty}^{q}\1\{X\le x\}\,dx=(q-X)_+.
\end{equation}
Therefore
\begin{equation}
    Z_{ES,N}(q)
    =\frac{1}{K_S}\sum_{t\in S}\left[
        \frac{1}{m}\sum_{i=1}^m(q-X_{t,i})_+
        -\E_{P_t}(q-X_t)_+
    \right].
\end{equation}
The conditional mean is zero, and conditional independence gives
\begin{equation}
    \Var[Z_{ES,N}(q)\mid\{P_t:t\in S\}]
    =\frac{1}{K_S^2m}\sum_{t\in S}\Var_{P_t}[(q-X_t)_+].
\end{equation}

\subsection{Proof of Theorem~\ref{thm:finite_budget_full}}
\label{subsec:proof_theorem}

We prove VaR first. Decompose
\begin{equation}
\widehat q^{DSI}_{\alpha,S,N}-q^\star
=\underbrace{(q_S-q^\star)}_{D_{2,q}:\,\text{trajectory}}
+\underbrace{(\widehat q^{DSI}_{\alpha,S,N}-q_S)}_{D_{1,q}:\,\text{sampling}}.
\end{equation}
For the trajectory term, Lemma~\ref{lem:functional_expansions} applied to $G=\bar F_S$ gives
\begin{equation}
    D_{2,q}
    =-\frac{\bar F_S(q^\star)-F^\star(q^\star)}{f^\star(q^\star)}+r^{traj}_{q,S}
    =-\frac{1}{f^\star(q^\star)}\frac{1}{K_S}\sum_{t\in S}U_t^q+r^{traj}_{q,S},
\end{equation}
where the deterministic remainder is negligible in the sense of Assumption~\ref{ass:two_stage_regularity}. Proposition~\ref{prop:checkpoint_averaging} then yields
\begin{equation}
    \E[D_{2,q}^2]
    =\frac{1}{f^\star(q^\star)^2}
    \left(b_q^2+\frac{\sigma_q^2}{K_{\mathrm{eff},q}(K_S,M)}\right)
    +o(L_q^{traj}).
\end{equation}

For the sampling term, Assumption~\ref{ass:stratified_bahadur} gives
\begin{align}
    D_{1,q} &= -\frac{Z_{q,N}}{\bar f_S(q_S)}+r^{sam}_{q,S,N}, \\
    Z_{q,N} &:= \widehat F_{S,N}(q_S)-\alpha.
\end{align}
Conditional on the trajectory, $\E[Z_{q,N}\mid\{P_t\}]=0$ because $\bar F_S(q_S)=\alpha$, and Proposition~\ref{prop:stratified_cdf} gives
\begin{equation}
    \Var[Z_{q,N}\mid\{P_t\}]=A^{\mathrm{mix}}_{q,N}(S).
\end{equation}
Therefore
\begin{equation}
    \E[D_{1,q}^2]
    =\E\left[\frac{A^{\mathrm{mix}}_{q,N}(S)}{\bar f_S(q_S)^2}\right]+o(N_S^{-1})
    =L_{q,N}^{sam}+o(N_S^{-1}).
\end{equation}
The leading sampling term has conditional mean zero, so
\begin{equation}
    \E[D_{1,q}D_{2,q}]
    =\E\!\big[D_{2,q}\,\E(r^{sam}_{q,S,N}\mid\{P_t\})\big].
\end{equation}
By Cauchy-Schwarz, Assumption~\ref{ass:two_stage_regularity}, and Appendix Assumption~\ref{ass:stratified_bahadur}, this cross term is negligible relative to $L_q^{traj}+L_{q,N}^{sam}$. Combining the three pieces proves the VaR expansion.

For ES, use the analogous decomposition
\begin{equation}
\widehat\ES^{DSI}_{\alpha,S,N}-\ES_\alpha(P^\star)
=\underbrace{(\ES_\alpha(\bar P_S)-\ES_\alpha(P^\star))}_{D_{2,ES}}
+\underbrace{(\widehat\ES^{DSI}_{\alpha,S,N}-\ES_\alpha(\bar P_S))}_{D_{1,ES}}.
\end{equation}
Lemma~\ref{lem:functional_expansions} gives
\begin{equation}
    D_{2,ES}
    =-\frac{1}{\alpha}\int_{-\infty}^{q^\star}(\bar F_S(x)-F^\star(x))\,dx+r^{traj}_{ES,S}
    =-\frac{1}{\alpha}\frac{1}{K_S}\sum_{t\in S}U_t^{ES}+r^{traj}_{ES,S},
\end{equation}
where the deterministic remainder is negligible in the sense of Assumption~\ref{ass:two_stage_regularity}. Proposition~\ref{prop:checkpoint_averaging} yields
\begin{equation}
    \E[D_{2,ES}^2]
    =\frac{1}{\alpha^2}\left(b_{ES}^2+\frac{\sigma_{ES}^2}{K_{\mathrm{eff},ES}(K_S,M)}\right)
    +o(L_{ES}^{traj}).
\end{equation}
For the sampling term, Assumption~\ref{ass:stratified_bahadur} gives
\begin{align}
    D_{1,ES} &= -\frac{1}{\alpha}Z_{ES,N}(q_S)+r^{sam}_{ES,S,N}, \\
    Z_{ES,N}(q_S) &:= \int_{-\infty}^{q_S}(\widehat F_{S,N}(x)-\bar F_S(x))\,dx.
\end{align}
By Proposition~\ref{prop:stratified_cdf}, conditional on the trajectory,
\begin{align}
    \E[Z_{ES,N}(q_S)\mid\{P_t\}] &= 0, \\
    \Var[Z_{ES,N}(q_S)\mid\{P_t\}] &= A^{\mathrm{mix}}_{ES,N}(S).
\end{align}
Thus
\begin{equation}
    \E[D_{1,ES}^2]
    =\frac{1}{\alpha^2}\E A^{\mathrm{mix}}_{ES,N}(S)+o(N_S^{-1}).
\end{equation}
The cross term is again negligible because the leading sampling term has conditional mean zero, the population remainder is controlled by Assumption~\ref{ass:two_stage_regularity}, and the sampling remainder is mean-square negligible by Appendix Assumption~\ref{ass:stratified_bahadur}. Combining trajectory, sampling, and cross terms proves the ES expansion.

\subsection{Proof of Corollary~\ref{cor:regimes}}
\label{subsec:proof_corollary}

The regimes follow directly from Theorem~\ref{thm:finite_budget_full}. If $b^2$ is small and $K_{\mathrm{eff}}$ grows, the checkpoint-fluctuation term $\sigma^2/K_{\mathrm{eff}}$ decreases, producing a variance-reducible regime. If $K_{\mathrm{eff}}\ll K_S$ because autocorrelation is persistent, the variance term does not decrease much, producing a correlation-limited regime. If $b^2$ dominates the other terms, increasing $K_{\mathrm{eff}}$ or $N$ cannot remove the dominant error, producing a bias-limited regime.

\subsection{Proof of Proposition~\ref{prop:failure_modes}}
\label{subsec:proof_failure_modes}

If $\rho(\ell M)=1$ for every $\ell=1,\ldots,K_S-1$, then
\begin{align}
    \kappa_\rho(K_S,M)
    &=1+2\sum_{\ell=1}^{K_S-1}\left(1-\frac{\ell}{K_S}\right) \nonumber\\
    &=1+2\left[(K_S-1)-\frac{1}{K_S}\frac{(K_S-1)K_S}{2}\right]
    =K_S.
\end{align}
Thus $K_{\mathrm{eff}}=K_S/K_S=1$.

For shared directional tail misspecification, suppose that for some $s\in\{-1,1\}$,
\begin{equation}
    s\{F_t(x)-F^\star(x)\}\ge a(x)
    \qquad\text{for every }t\in S.
\end{equation}
Then by linearity of the mixture CDF,
\begin{align}
    s\{\bar F_S(x)-F^\star(x)\}
    &=
    \frac{1}{K_S}\sum_{t\in S}s\{F_t(x)-F^\star(x)\} \nonumber\\
    &\ge
    \frac{1}{K_S}\sum_{t\in S}a(x)
    =a(x).
\end{align}
Thus the mixture inherits any common directional lower-tail error, including both undercoverage and overcoverage.

Finally, the bias floors follow by taking $K_{\mathrm{eff}}\to\infty$ and $N\to\infty$ in Theorem~\ref{thm:finite_budget_full}. The checkpoint-fluctuation and finite-sampling terms vanish, leaving $b_q^2/f^\star(q^\star)^2$ for VaR and $b_{ES}^2/\alpha^2$ for ES, up to negligible first-order remainders.

\section{Checkpoint Tail-Error Diagnostics}
\label{sec:checkpoint_diagnostics}

The following diagnostics support the regime interpretation in Section~\ref{sec:theory} and Section~\ref{sec:results}. For each strategy \(j\), let \(F_{t,j}\) and \(F^\star_j\) be the checkpoint and reference CDFs of the strategy-level PnL, and let \(q^\star_j=q_\alpha(P^\star_j)\). We compute the integrated ES checkpoint-error process \(U_{t,j}^{ES}:=\int_{-\infty}^{q^\star_j}(F_{t,j}(x)-F^\star_j(x))\,dx\) at \(\alpha=5\%\), before multiplication by \(1/\alpha^2\). Diagnostics are computed at the strategy level over the selected post-burn-in checkpoint sequence and then averaged over the \(J=65\) strategies. The purpose is not to introduce an additional benchmark, but to explain when checkpoint pooling should reduce error.

Table~\ref{tab:es_diagnostics} summarizes four quantities at \(K=200\) and \(M=10\). The bias floor \(B^2\) measures the squared persistent component of the trajectory error. The fluctuation variance \(\sigma^2\) measures the amount of checkpoint-level variation that can potentially be averaged. The sign-consensus statistic measures the average fraction of checkpoints sharing the dominant error sign. The effective checkpoint ratio \(K_{\mathrm{eff}}/K\) measures how much of the nominal checkpoint count remains after accounting for the strategy-specific autocorrelation of \(U_{t,j}^{ES}\). A favorable DSI regime has a small bias floor, non-negligible fluctuation variance, low sign consensus, and high \(K_{\mathrm{eff}}/K\).

\begin{table}[h]
\centering
\caption{Strategy-averaged integrated ES checkpoint-error diagnostics at $\alpha=5\%$, $K=200$, and $M=10$. Values for $B^2$ and $\sigma^2$ are in raw $U^{ES}$ units before multiplication by $1/\alpha^2$.}
\label{tab:es_diagnostics}
\begin{tabular}{lrrrr}
\toprule
Model & Bias floor $B^2$ & Fluctuation variance $\sigma^2$ & Sign consensus & $K_{\mathrm{eff}}/K$ \\
\midrule
DDPM    & $1.15\times 10^{-9}$ & $2.00\times 10^{-8}$ & $0.588$ & $0.953$ \\
TailGAN & $8.14\times 10^{-6}$ & $3.82\times 10^{-5}$ & $0.776$ & $0.149$ \\
WGAN    & $1.82\times 10^{-7}$ & $4.62\times 10^{-8}$ & $0.847$ & $0.647$ \\
\bottomrule
\end{tabular}
\end{table}

The diagnostics separate the three model trajectories. DDPM has a small bias floor and high \(K_{\mathrm{eff}}/K\), so its checkpoint fluctuations are averageable. TailGAN has substantial fluctuation variance, but its low \(K_{\mathrm{eff}}/K\) indicates strong autocorrelation, placing it in a correlation-limited regime. WGAN has relatively high sign consensus and a mixture-error term that remains large relative to its sampling term, indicating a shared-direction or bias-limited regime.

\begin{table}[h]
\centering
\caption{Strategy-averaged finite-budget ES integrated-tail decomposition at $\alpha=5\%$, $K=200$, $M=10$, and $N=1000$. Values are reported in raw integrated-tail units before multiplication by $1/\alpha^2$.}
\label{tab:finite_budget_diagnostics}
\begin{tabular}{lrrrr}
\toprule
Model & Mixture-error MSE & Sampling term $A_N$ & Predicted finite MSE & Observed fixed-budget MSE \\
\midrule
DDPM    & $1.229\times 10^{-9}$ & $9.370\times 10^{-9}$ & $1.060\times 10^{-8}$ & $1.074\times 10^{-8}$ \\
TailGAN & $8.142\times 10^{-6}$ & $3.127\times 10^{-8}$ & $8.173\times 10^{-6}$ & $8.141\times 10^{-6}$ \\
WGAN    & $1.824\times 10^{-7}$ & $6.495\times 10^{-9}$ & $1.889\times 10^{-7}$ & $1.888\times 10^{-7}$ \\
\bottomrule
\end{tabular}
\end{table}

Table~\ref{tab:finite_budget_diagnostics} compares the predicted finite-budget decomposition with the observed fixed-budget MSE, averaged over strategies and repeated fixed-budget draws. The mixture-error MSE is the population error of the checkpoint mixture \(\bar P_S\) relative to \(P^\star\), while \(A_N\) is the stratified sampling contribution under the fixed budget. For DDPM, the sampling term and the reduced mixture error are of comparable scale, so checkpoint pooling leads to a meaningful finite-budget improvement. For TailGAN, the predicted and observed errors are dominated by the mixture-error MSE, consistent with a correlation-limited trajectory. For WGAN, the mixture-error term also dominates the sampling term, consistent with a persistent bias floor. These diagnostics explain why DSI yields large gains for DDPM in Table~\ref{tab:merged_re}, while providing limited gains for the tested adversarial trajectories.

\begin{figure}
    \centering
    \includegraphics[width=0.9\textwidth]{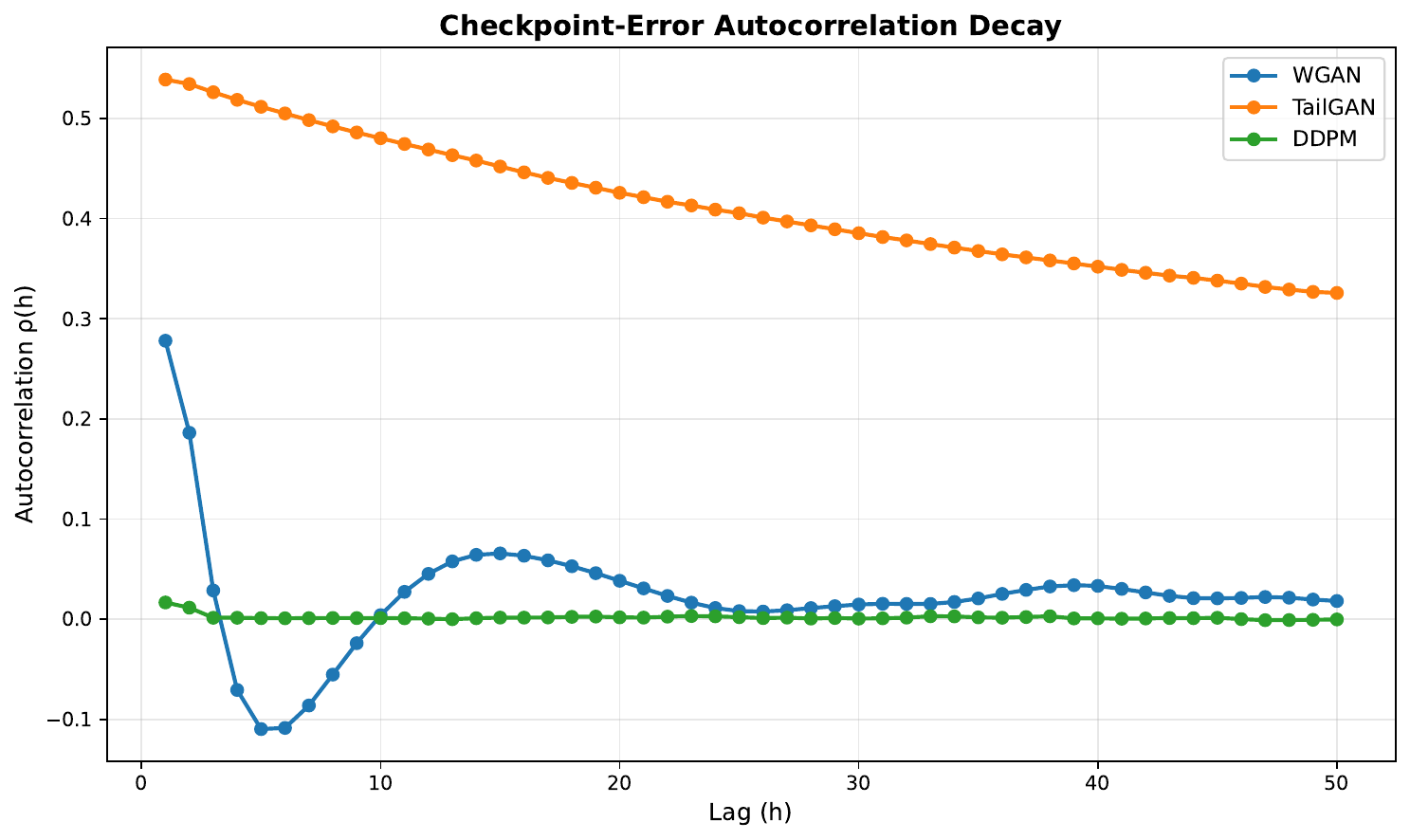}
    \caption{
    Autocorrelation decay of the post-burn-in checkpoint tail-error process. For each model, the curve reports the average lag-\(h\) autocorrelation across training checkpoints of a composite tail-error vector that concatenates strategy-level VaR error, ES error, and lower-tail CDF discrepancies relative to the reference distribution at \(\alpha=5\%\). Faster decay indicates that checkpoint-specific tail errors become less redundant as the stride increases, which increases the effective checkpoint count available to DSI. The DDPM curve decays most rapidly, consistent with the high \(K_{\mathrm{eff}}/K\) values in Table~\ref{tab:es_diagnostics}, TailGAN remains strongly correlated, while WGAN shows more persistent dependence than DDPM.
    }
    \label{fig:autocorr_decay}
\end{figure}

Figure~\ref{fig:autocorr_decay} provides a complementary visual summary of checkpoint dependence across the post-burn-in trajectory. Unlike Table~\ref{tab:es_diagnostics}, which reports strategy-averaged diagnostics for the integrated ES process \(U_{t,j}^{ES}\), this plot shows the lag-\(h\) autocorrelation of an aggregate checkpoint tail-error vector built from the strategy-level tail diagnostics used in the checkpoint analysis. The qualitative pattern is consistent with the regime interpretation above: DDPM exhibits faster decay toward zero, indicating more averageable checkpoint fluctuations, whereas TailGAN and WGAN retain stronger dependence across nearby checkpoints.

\section{Additional Implementation Details}
\label{sec:add_implementation}
In this section, we provide the formal procedural steps for the DSI framework and detail the experimental configurations.

\subsection{DSI SAMPLING PROCEDURE}
Algorithm \ref{alg:dsi} formalizes the test-time sample aggregation process introduced in Section \ref{subsec:ckp_ensembling} and \ref{subsec:ckp_stride}.

\begin{algorithm}[H]
\caption{Diachronic Sample Integration (DSI) Estimation}
\label{alg:dsi}
\KwIn{
Set of model checkpoints along a training trajectory $\{\theta_t\}_{t=1}^T$,
checkpoint count $K$,
temporal stride $M$,
total simulation budget $N$.
}
\KwOut{
Pooled synthetic dataset $\mathcal{X}$ representing samples from $\bar P_S$.
}

Identify start index $t_0$ after the burn-in phase\;
$S \leftarrow \{t_0, t_0 + M, t_0 + 2M, \dots, t_0 + (K-1)M\}$\;

Discard any indices in $S$ where $t > T$\;
$K_S \leftarrow |S|$\;

Set per-checkpoint sample size $m \leftarrow \lfloor N/K_S \rfloor$\;

$\mathcal{X} \leftarrow \emptyset$\;

\ForEach{$t \in S$}{
    Draw $m$ independent samples $\{X_{t,1}, \dots, X_{t,m}\} \sim P_{\theta_t}$\;
    $\mathcal{X} \leftarrow \mathcal{X} \cup \{X_{t,1}, \dots, X_{t,m}\}$\;
}

\Return $\mathcal{X}$.
\end{algorithm}

\paragraph{Checkpoint burn-in selection.} The checkpoint set $S(M,K)$ defined in Algorithm~\ref{alg:dsi} begins after a fixed burn-in period to avoid including under-trained models whose generative distributions deviate substantially from the target distribution.
To ensure reproducibility and prevent selection bias, we define the burn-in checkpoint $t_0$ deterministically as the midpoint of the training trajectory.

\begin{figure}
    \centering
    \includegraphics{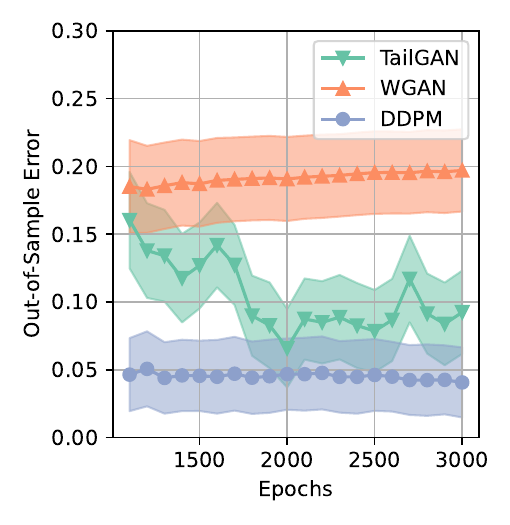}
    \caption{
    Post-burn-in out-of-sample error dynamics across training checkpoints. 
    The solid line shows average relative error at \(\alpha=5\%\), and the shaded region denotes one standard deviation. 
    DDPM maintains a low, stable error floor, TailGAN exhibits higher volatility, and WGAN shows an increasing error trend over later checkpoints.
    }
    \label{fig:training_graph}
\end{figure}

Specifically, models were trained for 3000 epochs, and checkpoints eligible for DSI were drawn uniformly from epochs 1000 onward. This choice coincides with the regime where the training loss and out-of-sample error have stabilized (Figure~\ref{fig:training_graph}), and does not rely on any tail-risk metric, validation risk estimate, or oracle information.

All reported DSI results use the same burn-in definition across models, datasets, and runs.

\subsection{Benchmark Trading Strategies}
\label{subsec:strategies}

We evaluate tail risk using a set of $J$ benchmark trading strategies adapted from \citet{Cont2025}. Each strategy acts as a deterministic functional mapping a multivariate asset price trajectory $\mathbf{S}_{1:T} = \{\mathbf{S}_t\}_{t=1}^T \in \mathbb{R}^{d \times T}$ (derived from the generated return sequence) to a univariate portfolio profit-and-loss (PnL) sequence. 

Crucially, the terminal PnL over the investment horizon physically instantiates the random variable $X$ introduced in Section~\ref{sec:problem}, upon which the statistical risk functionals $\VaR_\alpha(X)$ and $\ES_\alpha(X)$ are evaluated. To ensure strict causal evaluation, strategies are applied identically to empirical reference data and synthetic trajectories, utilizing only information available up to time $t$.

\subsubsection{Buy-and-Hold (Multivariate Dependence)}
The Buy-and-Hold strategy represents a static allocation baseline. Because it aggregates simultaneous price movements without dynamic trading, it explicitly tests the generative model's ability to preserve heavy-tailed multivariate dependence. 
Given a fixed portfolio weight vector $\mathbf{w} \in \mathbb{R}^d$ (derived from a predefined allocation matrix), the portfolio value $V_t$ at time $t$ is computed as the linear combination of the individual asset prices $S_{i,t}$:
\begin{equation}
    V_t = \mathbf{w}^\top \mathbf{S}_t = \sum_{i=1}^{d} w_i S_{i,t}.
\end{equation}
The strategy takes an initial position and holds it over the entire investment horizon. The terminal PnL is simply the difference between the final and initial portfolio values: $X = V_T - V_1$.

\subsubsection{Mean Reversion (Short-Term Temporal Dependence)}
The Mean Reversion (MR) strategy stress-tests the model's preservation of short-term negative autocorrelation, operating on the assumption that asset prices will eventually revert to a historical anchor. For a given asset, we define the anchoring mean $\mu_{\text{init}}$ as the average price over an initial lookback window of length $W$:
\begin{equation}
    \mu_{\text{init}} = \frac{1}{W} \sum_{k=1}^{W} S_k.
\end{equation}
At each subsequent active time step $t > W$, a normalized deviation score $z_t$ is computed to measure the divergence of the current price from this initial mean:
\begin{equation}
    z_t = \frac{S_t - \mu_{\text{init}}}{c},
\end{equation}
where $c > 0$ is a fixed empirical scaling factor (e.g., $c = 0.01$). Trading signals are generated dynamically based on empirical quantile thresholds. The strategy initiates a \textbf{Long} position when $z_t < \tau_{\text{lower}}$ (under the assumption that the asset is undervalued and will revert upward) and initiates a \textbf{Short} position when $z_t > \tau_{\text{upper}}$ (identifying overvaluation).

\subsubsection{Trend Following (Long-Term Temporal Dependence)}
The Trend Following (TF) strategy implements a Moving Average Crossover system to capitalize on market momentum, thereby probing the long-term positive autocorrelation of the generated paths. For a given asset, we compute two simple moving averages (SMA): a short-term average over a window of length $W$, and a long-term average over a window of length $2W$:
\begin{equation}
    \text{SMA}_{\text{short}, t} = \frac{1}{W} \sum_{k=0}^{W-1} S_{t-k}, \quad \text{SMA}_{\text{long}, t} = \frac{1}{2W} \sum_{k=0}^{2W-1} S_{t-k}.
\end{equation}
The trading signal is derived from the normalized divergence between these two temporal smoothing filters:
\begin{equation}
    z_t = \frac{\text{SMA}_{\text{short}, t} - \text{SMA}_{\text{long}, t}}{c}.
\end{equation}
Similar to the MR strategy, active market positions are determined by comparing $z_t$ against upper ($\tau_{\text{upper}}$) and lower ($\tau_{\text{lower}}$) thresholds. However, the trading logic is inverted to follow momentum: a \textbf{Long} position is initiated when the short-term average crosses significantly above the long-term average ($z_t > \tau_{\text{upper}}$), while a \textbf{Short} position is initiated when it crosses significantly below ($z_t < \tau_{\text{lower}}$).

\subsection{Setup of Parameters in the Synthetic Dataset}
\label{subsec:synthetic_setup}

We follow the synthetic data–generation procedure of \citet{Cont2025} exactly to ensure a fair and controlled comparison across methods.

For each time step $t \in \{1,\dots,T\}$, we first sample a latent innovation vector
\begin{equation}
\mathbf{u}_t = (u_{1,t}, \ldots, u_{5,t})^\top \sim \mathcal{N}(\mathbf{0}, \Sigma),
\end{equation}
where $\Sigma \in \mathbb{R}^{5\times5}$ is a covariance matrix.
In addition, we sample independent chi-square random variables
$v_{1,t} \sim \chi^2(\nu_1)$ and $v_{2,t} \sim \chi^2(\nu_2)$, which are independent of $\mathbf{u}_t$.

The five asset price increments are then generated according to
\begin{align*}
\Delta p_{1,t} &= u_{1,t}, \\
\Delta p_{2,t} &= \phi_1 \Delta p_{2,t-1} + u_{2,t}, \\
\Delta p_{3,t} &= \phi_2 \Delta p_{3,t-1} + u_{3,t}, \\
\Delta p_{4,t} &= \varepsilon_{4,t} = \sigma_{4,t}\eta_{1,t}, \\
\Delta p_{5,t} &= \varepsilon_{5,t} = \sigma_{5,t}\eta_{2,t},
\end{align*}
where the conditional variances evolve according to GARCH(1,1) dynamics:
\begin{align*}
\sigma^2_{4,t} &= \gamma_4 + \kappa_4 \varepsilon^2_{4,t-1} + \beta_4 \sigma^2_{4,t-1}, \\
\sigma^2_{5,t} &= \gamma_5 + \kappa_5 \varepsilon^2_{5,t-1} + \beta_5 \sigma^2_{5,t-1}.
\end{align*}
The standardized innovations are defined as
\begin{equation}
\eta_{1,t} = \frac{u_{4,t}}{\sqrt{v_{1,t}/\nu_1}}, 
\qquad
\eta_{2,t} = \frac{u_{5,t}}{\sqrt{v_{2,t}/\nu_2}},
\end{equation}
which induce heavy-tailed marginal behavior.

We set $T=100$, corresponding to the number of intraday observations within one trading day.
A correlation matrix $R$ is first generated by sampling each element independently from
$\mathrm{Unif}(0,1)$.
Annualized standard deviations $s_i$ are sampled uniformly from $[0.3, 0.5]$, and the covariance
matrix $\Sigma$ is constructed as
\begin{equation}
\Sigma_{ij} = \frac{s_i}{\sqrt{255T}} \, \frac{s_j}{\sqrt{255T}} \, R_{ij},
\qquad i,j = 1,\ldots,5.
\end{equation}

The remaining parameters are fixed as follows:
$\phi_1 = 0.5$, $\phi_2 = -0.15$,
$\nu_1 = 5$, $\nu_2 = 10$,
$\kappa_4, \kappa_5 \sim \mathrm{Unif}(0.08, 0.12)$,
$\beta_4, \beta_5 \sim \mathrm{Unif}(0.825, 0.875)$,
and
$\gamma_4, \gamma_5 \sim \mathrm{Unif}(0.03, 0.07)$.
Main results are reported for the risk level \(\alpha=0.05\), additional sensitivity results for \(\alpha\in\{0.01,0.05,0.10\}\) are provided in Appendix Table~\ref{tab:alpha_sensitivity}.

\begin{table*}[htbp]
\centering
\begin{tabular}{lcccccc}
\hline
 & \multicolumn{2}{c}{Static Buy-and-Hold} & \multicolumn{2}{c}{Mean-Reversion} & \multicolumn{2}{c}{Trend-Following} \\
\cline{2-7}
Asset & VaR & ES & VaR & ES & VaR & ES \\
\hline
AAPL & -0.366 & -0.591 & -0.307 & -0.491 & -0.382 & -0.527 \\
AMZN & -0.420 & -0.606 & -0.347 & -0.508 & -0.330 & -0.453 \\
GOOG & -0.377 & -0.572 & -0.340 & -0.510 & -0.347 & -0.462 \\
JPM  & -0.360 & -0.519 & -0.287 & -0.405 & -0.350 & -0.473 \\
QQQ  & -0.223 & -0.447 & -0.192 & -0.349 & -0.191 & -0.289 \\
\hline
\end{tabular}
\caption{Empirical Value-at-Risk (VaR) and Expected Shortfall (ES) on market training data for $\alpha=5\%$.}
\label{tab:market_data_summary}
\end{table*}
\subsection{Diffusion Parameters}
\label{app:default_diffusion}
Following the Denoising Diffusion Probabilistic Model (DDPM) framework~\citet{Ho2020}. The construction consists of a forward diffusion process that gradually adds noise to the data and a reverse process parameterized by a neural network to denoise it.

\paragraph{Diffusion Process.}
We define a forward process $q(\mathbf{x}_t | \mathbf{x}_{t-1})$ which adds Gaussian noise to the data $\mathbf{x}_0$ over $T_{\mathrm{diff}}=1000$ timesteps according to a variance schedule $\beta_t$:
\begin{equation}
    q(\mathbf{x}_t | \mathbf{x}_{t-1}) = \mathcal{N}(\mathbf{x}_t; \sqrt{1 - \beta_t} \mathbf{x}_{t-1}, \beta_t \mathbf{I}).
\end{equation}
We utilize a linear noise schedule where $\beta_t$ increases linearly from $\beta_1 = 10^{-4}$ to $\beta_{T_{\mathrm{diff}}} = 0.02$.

\paragraph{Network Architecture.}
The reverse process is modeled by a 1D U-Net architecture, denoted as $\epsilon_\theta$, which takes the noisy input $\mathbf{x}_t \in \mathbb{R}^{d \times L}$ (where $d$ is the number of assets and $L$ is the time horizon) and the timestep $t$ to predict the added noise. The network consists of:
\begin{enumerate}
    \item \textbf{Time Embedding:} A sinusoidal position embedding followed by a Multi-Layer Perceptron (MLP) with SiLU activation, projecting the discrete timestep $t$ into a $256$-dimensional embedding vector.
    \item \textbf{Encoder:} A series of 1D residual blocks and downsampling layers. Each residual block comprises two 1D convolutional layers with Group Normalization (groups=8), SiLU activations, and a time embedding projection that is added to the intermediate feature map. Downsampling is performed via strided 1D convolutions.
    \item \textbf{Bottleneck:} A sequence of residual blocks processing the latent representation at the lowest resolution.
    \item \textbf{Decoder:} A symmetric path of upsampling layers (transposed 1D convolutions) and residual blocks. Skip connections concatenate feature maps from the encoder to the decoder to preserve high-frequency details.
\end{enumerate}
The model was trained to minimize the mean squared error (MSE) between the true noise $\epsilon$ and the predicted noise $\epsilon_\theta(\mathbf{x}_t, t)$:
\begin{equation}
    \mathcal{L} = \mathbb{E}_{t, \mathbf{x}_0, \epsilon} \left[ \| \epsilon - \epsilon_\theta(\sqrt{\bar{\alpha}_t}\mathbf{x}_0 + \sqrt{1-\bar{\alpha}_t}\epsilon, t) \|^2 \right],
\end{equation}
where $\bar{\alpha}_t = \prod_{s=1}^t (1 - \beta_s)$. The network uses a base channel dimension of $64$ and is optimized using AdamW with a learning rate of $10^{-5}$.

\section{Additional Experimental Results}
\label{sec:additional_results}
This section provides supplementary empirical results that support and contextualize the main findings in Section~\ref{sec:results}.

\begin{table}[t]
\centering
\small
\setlength{\tabcolsep}{6pt}
\caption{
Sensitivity of tail-risk estimation error across tail levels under the fixed budget \(N=1000\).
The reported metric is average relative error of VaR and ES in percent. Lower is better.
}
\label{tab:alpha_sensitivity}
\begin{tabular}{lcccccc}
\toprule
& \multicolumn{3}{c}{Synthetic} & \multicolumn{3}{c}{Market} \\
\cmidrule(lr){2-4} \cmidrule(lr){5-7}
Method
& \(\alpha=0.01\)
& \(\alpha=0.05\)
& \(\alpha=0.10\)
& \(\alpha=0.01\)
& \(\alpha=0.05\)
& \(\alpha=0.10\) \\
\midrule
TailGAN
& 6.9 $\pm$ 3.1 & 5.1 $\pm$ 1.4 & 4.5 $\pm$ 1.2
& 17.8 $\pm$ 3.8 & 12.1 $\pm$ 1.3 & 10.9 $\pm$ 1.1 \\
DDPM (EMA)
& 9.1 $\pm$ 3.6 & 7.7 $\pm$ 2.5 & 6.1 $\pm$ 1.9
& 27.3 $\pm$ 6.1 & 22.3 $\pm$ 4.3 & 19.2 $\pm$ 3.5 \\
DDPM (DSI)
& \textbf{4.6 $\pm$ 0.3} & \textbf{3.7 $\pm$ 0.12} & \textbf{3.4 $\pm$ 0.10}
& \textbf{12.2 $\pm$ 2.1} & \textbf{10.8 $\pm$ 1.2} & \textbf{8.9 $\pm$ 0.9} \\
\bottomrule
\end{tabular}
\end{table}

\begin{table*}
\centering
\begin{tabular}{@{}l|ccccccc@{}}
\toprule
& \multicolumn{7}{c}{\textbf{Number of Checkpoints}} \\ \cmidrule(l){2-8} 
\textbf{Stride} & \textbf{5} & \textbf{10} & \textbf{20} & \textbf{50} & \textbf{100} & \textbf{150} & \textbf{200} \\ \midrule
\textbf{1} & 4.80 $\pm$ 0.46 & 4.74 $\pm$ 0.34 & 4.39 $\pm$ 0.17 & 3.84 $\pm$ 0.11 & 3.90 $\pm$ 0.15 & 4.00 $\pm$ 0.16 & 3.83 $\pm$ 0.17 \\
\textbf{2} & 4.91 $\pm$ 0.63 & 4.23 $\pm$ 0.41 & 3.73 $\pm$ 0.21 & 3.85 $\pm$ 0.22 & 3.80 $\pm$ 0.15 & 3.77 $\pm$ 0.12 & 3.94 $\pm$ 0.13 \\
\textbf{5} & 4.76 $\pm$ 0.62 & 4.47 $\pm$ 0.38 & 3.92 $\pm$ 0.25 & 3.84 $\pm$ 0.12 & 3.81 $\pm$ 0.19 & 3.85 $\pm$ 0.10 & 3.71 $\pm$ 0.14 \\
\textbf{10} & 4.71 $\pm$ 0.79 & 4.19 $\pm$ 0.25 & 4.06 $\pm$ 0.18 & 4.10 $\pm$ 0.27 & 3.66 $\pm$ 0.12 & 3.69 $\pm$ 0.19 & 3.82 $\pm$ 0.12 \\
\textbf{20} & 4.68 $\pm$ 0.49 & 4.24 $\pm$ 0.52 & 4.09 $\pm$ 0.31 & 3.95 $\pm$ 0.25 & 3.94 $\pm$ 0.16 & 4.05 $\pm$ 0.13 & 3.81 $\pm$ 0.12 \\ \bottomrule
\end{tabular}
\caption{Full sensitivity analysis results: the reported metric is the average RE of VaR and ES in \% at alpha = 5\%.}
\label{tab:full_sensitivity_results}
\end{table*}

\begin{figure*}
    \centering
    \includegraphics[width=\textwidth]{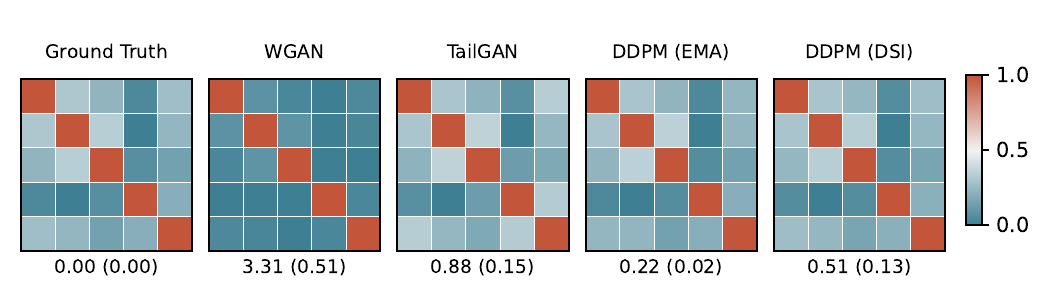}
    \caption{
    Multivariate correlations of the price increments in the Synthetic data and from different trained models. Numbers below each heatmap: mean and standard deviation (in parentheses) of the sum of the absolute difference between the correlation coefficients computed with all training samples and 1,000 generated samples.
    }
    \label{fig:synthetic_correlation}
\end{figure*}

\begin{figure*}
    \centering
    \includegraphics[width=\textwidth]{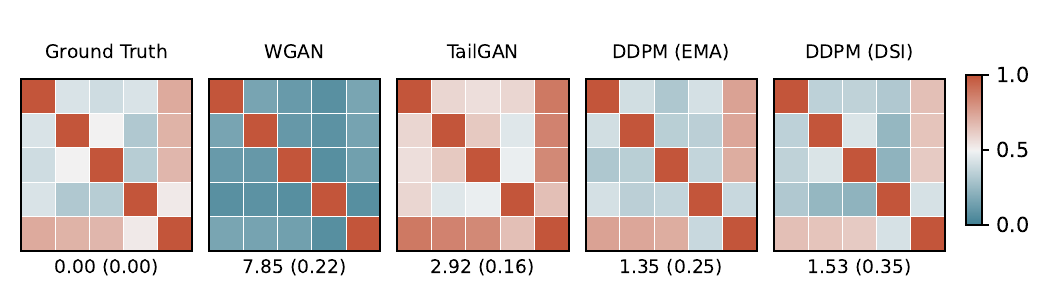}
    \caption{
    Cross-asset correlations of the price increments in the Market data and from different trained models. Numbers below each heatmap: mean and standard deviation (in parentheses) of the sum of the absolute difference between the correlation coefficients computed with all training samples and 1,000 generated samples.
    }
    \label{fig:market_correlation}
\end{figure*}

\clearpage

\begin{figure*}
    \centering
    \includegraphics[width=\textwidth]{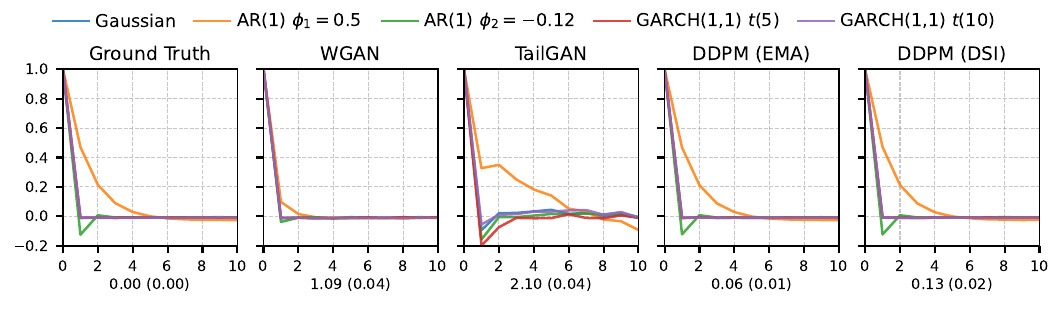}
    \caption{Auto-correlations of the price increments from different trained models for Synthetic data. Numbers below each graph: mean and standard deviation (in parentheses) of the sum of the absolute element-wise difference between auto-correlation coefficients computed with all training samples and 1,000 generated samples.
    }
    \label{fig:synthetic_autocorr}
\end{figure*}

\begin{figure*}
    \centering
    \includegraphics[width=\textwidth]{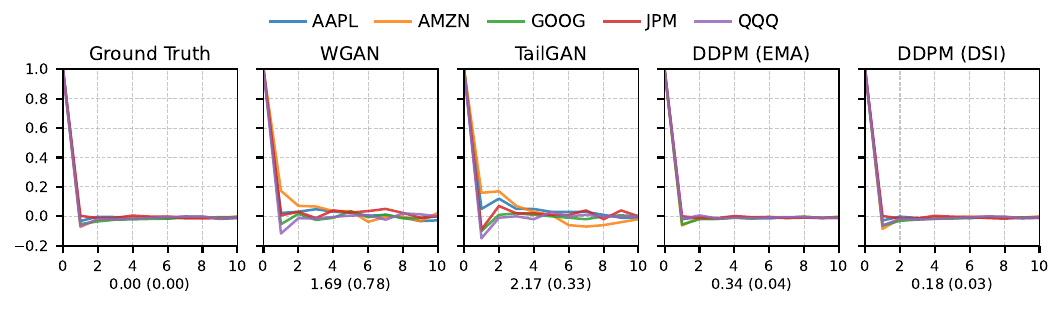}
    \caption{Auto-correlations of the price increments from different trained models for Market data. Numbers below each graph: mean and standard deviation (in parentheses) of the sum of the absolute element-wise difference between auto-correlation coefficients computed with all training samples and 1,000 generated samples.}
    \label{fig:market_autocorr}
\end{figure*}

\begin{figure*}[p]
    \centering
    \includegraphics[width=0.9\textwidth]{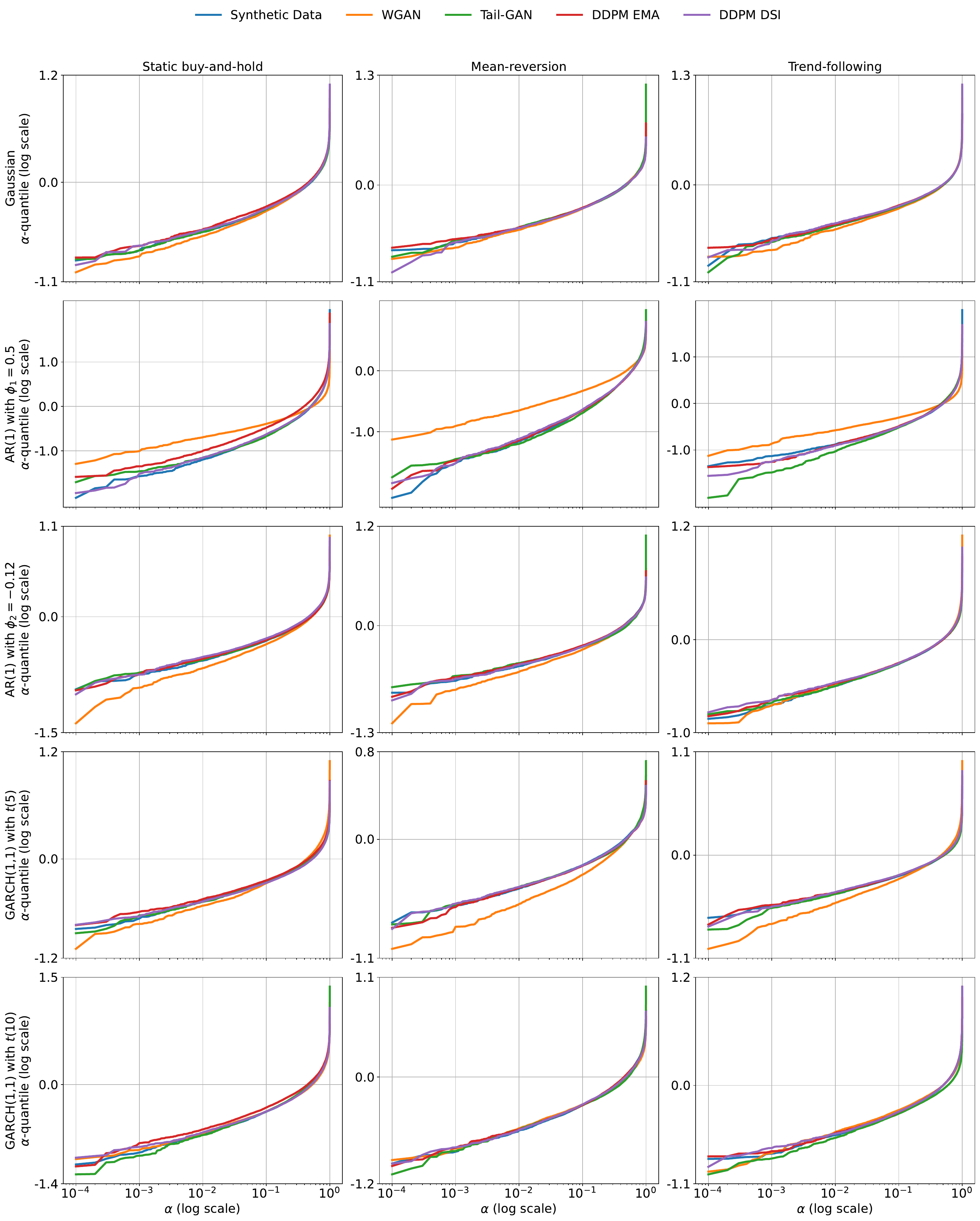}
    \caption{Tail behavior via the synthetic rank-frequency distribution of the strategy PnL. The columns represent the strategy types.}
    \label{fig:synthetic_rankfreq}
\end{figure*}

\begin{figure*}[t]
    \centering
    \includegraphics[width=0.9\textwidth]{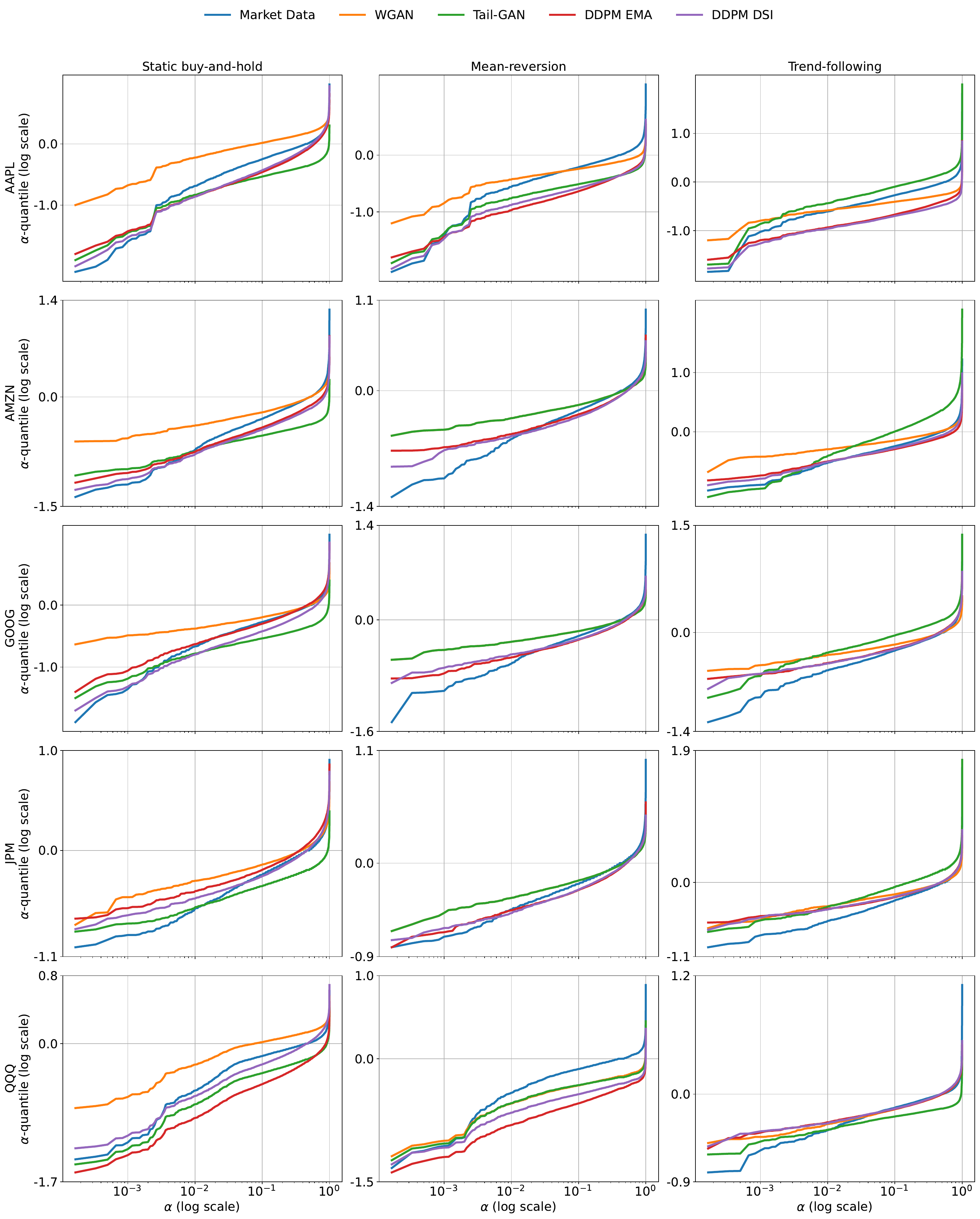}
    \caption{Tail behavior via the market rank-frequency distribution of the strategy PnL. The columns represent the strategy types.}
    \label{fig:market_rankfreq}
\end{figure*}

\end{document}